\documentclass[11pt]{article}
\usepackage[final]{acl}

\usepackage{times}
\usepackage{latexsym}
\usepackage{float}
\usepackage{microtype}
\usepackage{inconsolata}

\usepackage{amsmath}
\usepackage{amssymb}
\usepackage{graphicx}
\usepackage{booktabs}
\usepackage{multirow}
\usepackage{xcolor}
\usepackage{url}
\usepackage{pgfplots}
\pgfplotsset{compat=1.17}
\usepackage{cleveref}
\usepackage[T1]{fontenc}
\usepackage[utf8]{inputenc}
\usepackage{tcolorbox}
\tcbuselibrary{listings,skins,breakable}

\usepackage{algorithm}
\usepackage{algpseudocode}
\usepgfplotslibrary{groupplots} 

\usepackage[table]{xcolor}
\definecolor{lreGreen}{HTML}{1A7F37}
\definecolor{fifoOrange}{HTML}{BC4C00}
\definecolor{aconRed}{HTML}{CF222E}
\definecolor{boxBlue}{HTML}{0969DA}
\definecolor{snipBg}{HTML}{F6F8FA}
\definecolor{mintgreen}{HTML}{DDF5E3}
\definecolor{mintFrame}{HTML}{86c395}

\lstdefinestyle{snip}{basicstyle=\ttfamily\footnotesize, breaklines=true,
  breakatwhitespace=true, backgroundcolor=\color{snipBg}, frame=none,
  columns=fullflexible, showstringspaces=false, aboveskip=2pt, belowskip=2pt}
\newtcolorbox{qualbox}[2][]{enhanced, breakable, colback=white,
  colframe=mintFrame, coltitle=white, fonttitle=\bfseries, title={#2},
  boxrule=0.8pt, left=6pt, right=6pt, top=4pt, bottom=4pt, #1}

\usepackage{pifont}
\newcommand{\method}{LRE}

\title{\textbf{Learning What Not to Forget: Long-Horizon Agent Memory from a Few Kilobytes of Learning}}
\author{}
\date{}

\author{
  Nusrat Jahan Lia \\
  Institute of Information Technology \\
  University of Dhaka, Dhaka, Bangladesh \\
  \texttt{bsse1306@iit.du.ac.bd} \And
  Aritra Mazumder \\
  University of Utah \\
  Utah, USA \\
  \texttt{aritra.mazumder@utah.edu}
}
\begin{document}
\maketitle

\begin{abstract}
Long-running language-model systems accumulate interaction history that outgrows the context window, so they must continually evict. When an eviction policy drops a task-critical detail, for example an access token issued at login or a path the next call needs, the action fails. We present \method{} (Learned Relevance Eviction), a kilobyte-scale, CPU-only, language-model-free scorer that learns which units of history are task-critical and keeps them by verbatim extraction. Under a matched-budget comparison, in our experiment, no baseline dominates \method{} on the accuracy-cost plane. On agents, \method{} recovers $93\%$ of the aggregate accuracy of keeping the entire history ($41.1$ vs.\ $44.0$) and exceeds it by $27\%$ on the simplest tasks, while requiring zero compressor calls and cutting the worst-case peak prompt by $52\%$. A controlled study trace shows \method{} completes tasks where the others loop, finishing one such task in $37\%$ fewer calls than keeping everything and solving $14$ tasks where no other run policy does. On conversational memory, \method{} outranks dense and token-pruning encoders at zero neural cost while being $295$--$1569\times$ smaller in size. In downstream evaluation, \method{} gives the best budgeted answer quality on LoCoMo reading $68\%$ fewer tokens. Its supervision can also be annotation-free: training only on the system's own behavior recovers $95\%$ of the supervised scorer's effectiveness. We argue that, because memory eviction in LLM agents is a fidelity problem, it requires a deployable proactive policy where the future query is unavailable and exact state is decisive, and that cheap learned relevance can be sufficient.\footnote{https://github.com/NusRAT-LiA/LRE}
\end{abstract}

\begin{figure}[t]
  \centering
  \includegraphics[width=\linewidth]{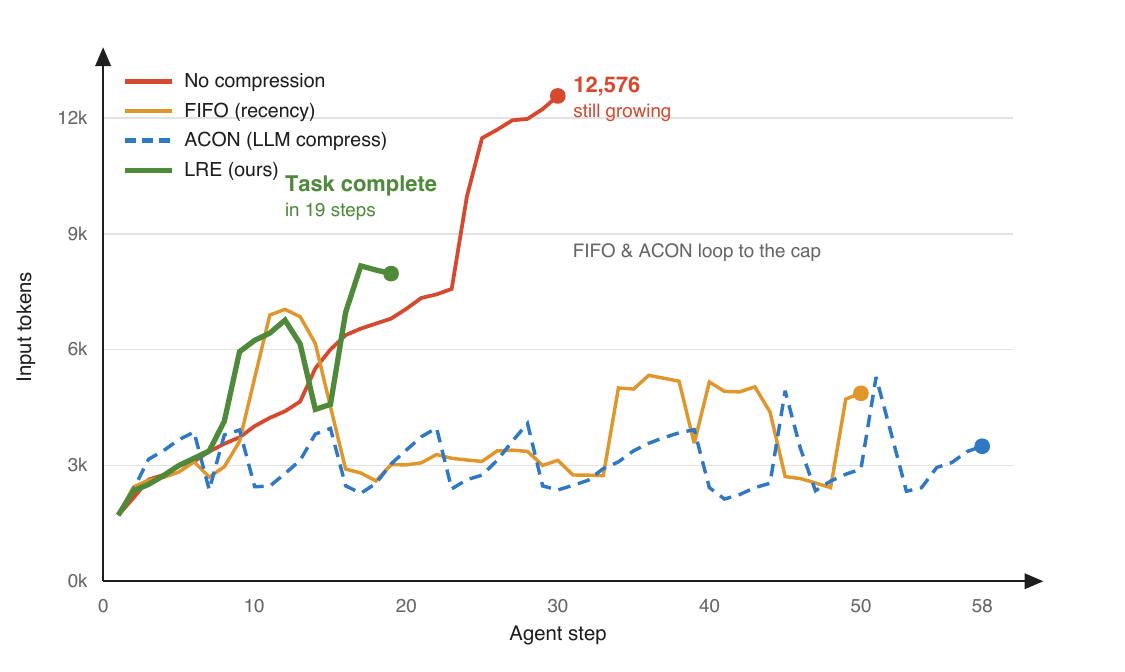}
  \caption{\textbf{How an agent evicts decides whether it finishes.} Context tokens per generation on one AppWorld task (\texttt{d194965\_2}) under four policies at a $2048$-token budget. No-compression grows unbounded (${\sim}12.6$k tokens); FIFO and LLM compression (ACON) stay bounded but lose the task-critical content. ACON does 49 llm + 8 compressor calls. \method{} keeps the relevant turns \emph{verbatim} with a few kilobyte CPU scorer and completes in $19$ calls, 37\% fewer than no-compression’s 30.}
  \label{fig:growth}
\end{figure}

\section{Introduction}
\label{sec:intro}

\begin{figure*}[t]
  \centering
  \includegraphics[width=\linewidth]{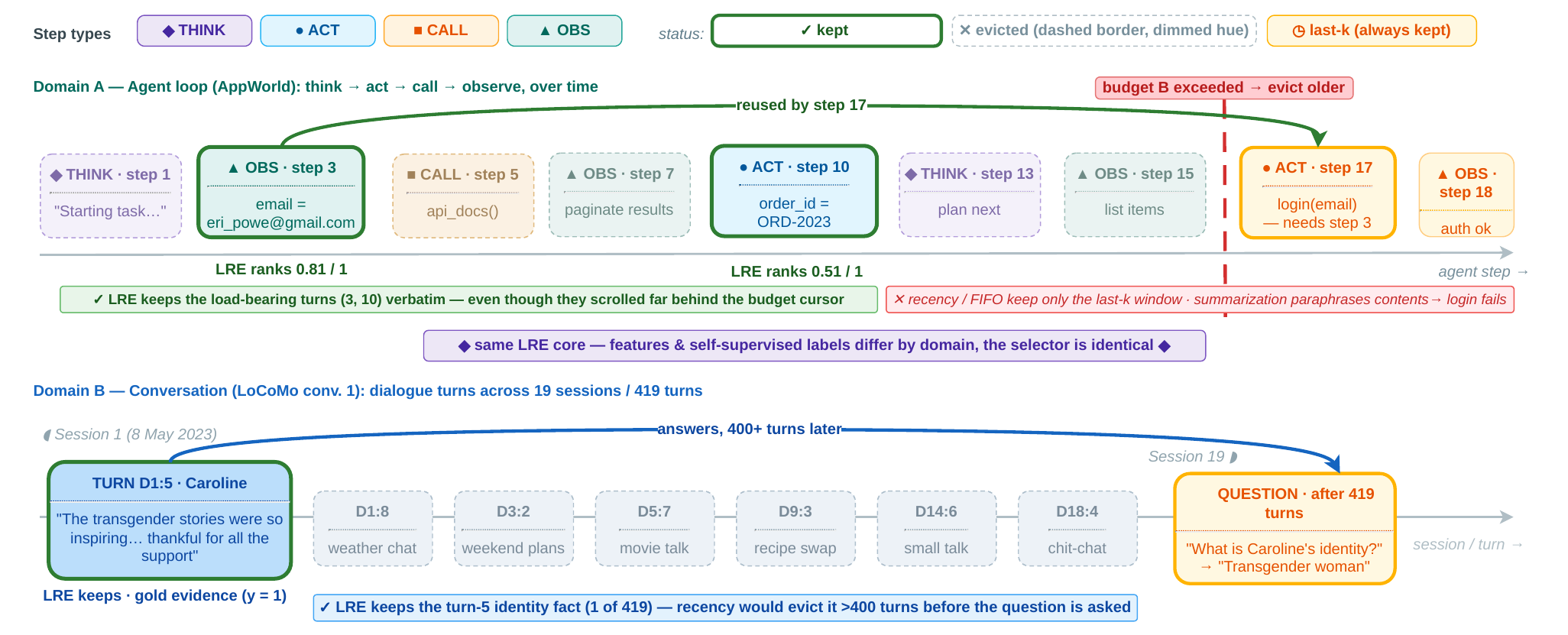}
    \caption{
  \textbf{Top (agent, AppWorld).} Across a \texttt{think/act/call/observe} loop, a credential
  observed at step 3 and an identifier produced at step 10 scroll far behind the budget cursor. \method{} ranks them task-critical and keeps them verbatim, so the login at step 17 succeeds. \textbf{Bottom (conversation, LoCoMo conv).} An identity fact stated at turn 5 of a  $419$-turn, $19$-session history answers a question asked more than $400$ turns later. \method{}  keeps that evidence turn.}
  \label{fig:method}
\end{figure*}

Large language models act as long-horizon agents and assistants that
plan and act over extended interactions. A tool-using agent
accumulates hundreds of action and observation steps, and a conversational
assistant accumulates dialogue across dozens of sessions. In both cases the history outgrows the context window. The memory and compute cost of attention grows with context length, making long histories expensive to serve. Overly
long contexts dilute relevant information, distract the model with stale detail,
and degrade decision quality \citep{liu2024lost}. The system must therefore
\textbf{evict}: choose, within a token budget, the subset of history that preserves
downstream quality. When eviction drops a task-critical detail, the action fails even if the context is accurate.

Eviction is different from retrieval, in a way that rules out the standard solution. A
retriever runs at query time and conditions on the query \citep{karpukhin2020dense,lewis2020retrieval}; an eviction policy which runs at write time must decide what to keep \emph{before} the query or action that needs it exists. The decision is proactive and query-blind, which excludes query-conditioned retrieval and explains why deployed systems fall back on reactive heuristics such as recency or decay \citep{xiao2023streaming,zhong2024memorybank}. One way is calling a language model to compress history \citep{jiang2023llmlingua,pan2024llmlingua2,kang2025acon} which comes with non-deterministic behavior and paraphrasing challenges. Some methods learn what to retain rather than relying on heuristics \citep{chung2024selection,bui2025trimkv}, but score each unit with a forward pass or distilled gates tied to specific layers and heads, so retention still costs a neural evaluation and stays coupled to the host model.

We argue the right primitive for eviction is a learned \emph{relevance} estimate:
cheap enough to run at every write, and faithful enough to keep verbatim what
matters. We instantiate it as \method{} (Learned Relevance Eviction), a scorer over cheap causal features that assigns each unit a keep-probability, retaining the highest-scoring units within budget via verbatim selection. The scorer can be trained on already logged data, on the order of kilobytes ($295$--$1569\times$ smaller than the neural encoders it outranks; \Cref{tab:scorer-size}), runs on CPU with no neural forward pass
and no language-model call at compression time, and scores $1{,}181$ units per second
where a dense encoder manages $103$ and an LLM token-pruner $36$ ($11.5\times$ and
$32.8\times$ slower respectively; Table~\ref{tab:conv-eff}).

On agents at a $2048$-token budget, \method{} completes tasks in 19 steps where the compressor loops to 49 (\Cref{fig:growth}) and reaches $41.1\%$ task success against full history's $44.0\%$ while needing zero compressor calls (\Cref{tab:appworld-main}). The scorer is external to the host model and transfers across model families without retraining (\Cref{app:crossmodel}). On conversational memory the scorer ranks evidence at $0.829$ macro-AUC on LoCoMo and $0.708$ on LongMemEval$_S$ without ever seeing the question, above dense ($0.643$) and token-pruning ($0.402$) encoders that each spend a forward pass (\Cref{tab:convqa}). A natural question is that supervision for ``relevance'' requires expensive annotation. We show it does not: a self-supervised label derived from the system’s own later behavior recovers about $95\%$ of the supervised model’s signal beyond random ordering (\Cref{tab:selfsup}). The signal a deployed system needs to learn proactive retention, is already present in its logs.

We present:

\begin{itemize}
  \item \textbf{\method{}, a learned query-blind eviction policy.} We present a kilobyte-scale, CPU-only, language-model call-free scorer method that learns which units of history are task-critical and keeps them by verbatim extraction, with one design serving both  conversational memory and long-horizon agents.

  \item \textbf{Pareto efficiency in both domains.} Under matched budgets, \method{} is an efficient and cheaper baseline that preserves or exceeds performance gains. On agent tasks it recovers $93\%$ of the aggregate success rate of retaining the entire history ($41.1$ vs.\ $44.0$) and exceeds it on Easy tasks, while cutting the worst-case peak prompt by ${\sim}52\%$ and transferring zero-shot across model families (details in \Cref{app:appworld-extra}, \Cref{app:crossmodel}); on conversational memory, in retaining useful content, it outranks dense encoders without a forward pass of its own (\Cref{tab:convqa}).

  \item \textbf{Fidelity, not just retrieval, is what eviction needs.} A controlled study traces \method{}'s agent advantage to verbatim fidelity: extraction keeps the task-critical context a task needs in the future while compression or recency loses them, leaving fourteen tasks that could be solved by \method{} alone (\Cref{sec:results-case}). 

\end{itemize}


\section{Related Work}
\label{sec:related-work}

\paragraph{Memory architectures for conversational and agentic LLMs.}
LLM memory systems decide both what to \emph{write} to long-term store and what to \emph{keep} active. MemGPT \citep{packer2023memgpt} borrows virtual-memory paging to move content between context and external storage, while MemoryBank \citep{zhong2024memorybank} folds both decisions into a time-based exponential decay model. On the write side, prior works have organized history into episodes, reflections, or linked notes that are periodically consolidated \citep{park2023generative,xu2026mem,chhikara2025mem0,nan2025nemori}, and event-centric variants decompose dialogue into propositions to keep recall non-lossy \citep{zhou2025emem}. These mechanisms improve what is stored, but retention \emph{within} a tier defaults to recency, decay, or a further LLM call.

\paragraph{Reactive eviction under a fixed budget.}
A parallel line of work evicts directly from a size-capped buffer using cheap,
query-blind signals. StreamingLLM \citep{xiao2023streaming} retains a few initial
attention-sink tokens alongside a recency window; H\textsubscript{2}O \cite{zhang2023h2o} keeps the tokens that have received the most attention so far, scoring a unit by how it has already been attended to rather than by whether a later step may need it, and operating on attention statistics at the token level inside a forward pass.

\paragraph{Learned retention.}
Recent work \emph{learns} what to keep rather than
relying on attention heuristics. Selection-p \citep{chung2024selection} learns
self-supervised, task-agnostic token retention, and TRIM-KV
\citep{bui2025trimkv} learns a per-token retention gate at creation time that
decays over the trajectory and evicts under a budget. Both confirm that learned
retention can beat hand-designed decay. Their signal is produced inside the
model, by forward passes or distilled gates tied to specific layers and heads,
so each unit still costs a neural evaluation. We address whether such predictive principle holds for \emph{content units} at \emph{zero} added inference cost, with a scorer external to and independent of the generating model.

\paragraph{LLM-based compression and its cost.}
At the expensive end, the model itself does the scoring. LLMLingua \citep{jiang2023llmlingua} ranks tokens by perplexity under a small language model and prunes iteratively under a budget controller; LLMLingua-2 \citep{pan2024llmlingua2} instead distills GPT-4's extractive compression capabilities into a trained, bidirectional token classifier. Both optimize for task-agnostic compression rather than for any downstream task. In the agentic setting,
context-management methods issue an LLM call to summarize or curate the
accumulated history \citep{kang2025acon}, adding latency on the critical path.
Such summaries are typically conditioned on the current task, the query-aware
advantage makes them a strong upper bound but not a deployable peer for a
policy that can decide before the task context exists.

\paragraph{Related directions.}
Adjacent work studies proactive computation prediction
\citep{pan2026kvflow,zheng2026efficient,ye2025agentfold},
systems-level context management \citep{kwon2023efficient},
and annotation-free supervision
\citep{andrychowicz2017hindsight,izacard2021unsupervised}; we discuss
these connections further in Appendix~\ref{app:extended-related}.

Along the cost axis, prior work spans zero neural cost (recency, MemoryBank,
StreamingLLM), one neural pass per unit (dense and lexical salience,
H\textsubscript{2}O, Selection-p, TRIM-KV), and one or more LLM calls per
decision (LLMLingua, ACON, AgentFold). Along deployability, it splits into
query-blind and query-aware scoring, and along timing, into reactive and proactive eviction. We therefore bridge that gap with a \emph{learned}, \emph{query-blind}, \emph{extractive} scorer that anticipates which past content a future query will need, at \emph{zero} added inference cost and independent of the host model, evaluated on both agentic \citep{trivedi2024appworld} and conversational \citep{wu2024longmemeval,maharana2024loco} memory with downstream gains. 


\section{Methodology}
\label{sec:method}

\subsection{Framework: Score, Retain, Evict}
\label{sec:method-framework}

A long-running system holds a history $H=[u_1,\dots,u_t]$ of units, where a unit is a dialogue turn or
an agent action-observation step. \method{} assigns each unit a relevance score, retains the highest-scoring units under a budget, and evicts the rest.
Let
\[
\phi : (u_i,\,u_{\le i}) \to \mathbb{R}^d
\]
be a feature map whose value at unit $i$ depends only on the prefix $u_{\le i}$, never on
$u_{>i}$ and never on any test-time query. This causality invariant makes the inference
representation identical to the training representation which rules out evaluation leakage at the
feature level.
The scorer is
\[
p_i \;=\; f_\theta(u_i) \;=\; \sigma\!\big(\theta^\top \phi(u_i, u_{\le i})\big),
\]
an L2-regularized logistic regression with parameters $\theta \in \mathbb{R}^d$ fit on logged data, where $p_i$ denotes the estimated relevance of unit $u_i$. 
Given per-unit costs $c_i$, a budget $B$ on the evictable older history, and an always-retained
recent window $R_k\subseteq H$ that is protected and concatenated on top (along with the system
prompt and task), \method{} selects
\begin{equation}
\begin{aligned}
S^\star \;&=\; \arg\max_{S \subseteq H \setminus R_k}\;\sum_{u_i \in S} p_i, \\
\text{s.t.}\;&\sum_{u_i \in S} c_i \,\le\, B,
\end{aligned}
\label{eq:lre-knapsack}
\end{equation}
\[
\hat H \;=\; \operatorname{sort}_{\text{time}}(S^\star \cup R_k).
\]
The cap $B$ applies to $S^\star$ only; $R_k$, the system prompt, and the task are emitted unchanged,
so the total context size is $\mathrm{tokens}(\text{system})+\mathrm{tokens}(\text{task})+
\sum_{u \in S^\star} c_u + \sum_{u \in R_k} c_u$ in the worst case.
Equation~\eqref{eq:lre-knapsack} is a $0$/$1$ knapsack. Algorithm~\ref{alg:lre} uses the standard greedy value-density heuristic; when $c_i \equiv 1$ this reduces to the exact top-$K$ optimum.

\begin{figure}[t]
\centering
\fbox{%
\begin{minipage}{0.97\columnwidth}
\begin{algorithmic}[1]
\Require $\{u_i\}_{i=1}^t$, $\{p_i\}$, $\{c_i\}$, $B$, $R_k$
\State $S \gets R_k$;\ \ $\mathrm{used} \gets 0$
        \Comment{$\mathrm{used}$ tracks only older selected units; $R_k$ is uncapped}
\For{$i$ in order of decreasing $p_i / c_i$}
    \If{$u_i \in R_k$} \textbf{continue} \EndIf
    \If{$\mathrm{used} + c_i \le B$}
        \State $S \gets S \cup \{u_i\}$;\ \ $\mathrm{used} \gets \mathrm{used} + c_i$
    \EndIf
\EndFor
\State \Return $\operatorname{sort}_{\mathrm{time}}(S)$
\end{algorithmic}
\end{minipage}}
\caption{\method{}: budgeted extractive eviction. The cap $B$ applies to the older selected units
$S^\star$; the recent window $R_k$ (and the system prompt and task, omitted here) are appended
without competing for $B$. Older units are added in decreasing value density $p_i / c_i$, skipping
any that overflow. The two domains instantiate $(c_i, B, R_k)$ as in
Table~\ref{tab:lre-instantiations}.}
\label{alg:lre}
\end{figure}

\begin{figure}[t]\centering
\input{figures/fig_label_extraction}
\caption{The self-supervised label in each domain, on one real example. The agent label fires on an
identifier reused by $\ge 3$ later steps; the conversational label fires on a turn whose content is
answered later (and coincides with the human gold-evidence label). Choice of values are justified in \Cref{app:justify_label}.}
\label{fig:label-extraction}
\end{figure}

\begin{table}[t]\centering\small
\begin{tabular}{@{}lll@{}}
\toprule
\textbf{Symbol} & \textbf{Domain A} & \textbf{Domain B} \\
\midrule
$c_i$ & $\mathrm{tokens}(u_i)$ & $1$ (unit count)  \\
$B$   & $2048$ tokens & $\lceil b\,n\rceil$,\ $b\in\{0.1,0.2,0.3, 0.4\}$ \\
$R_k$ & last $5$ units & $\emptyset$ (no last-$k$ carve-out) \\
\bottomrule
\end{tabular}
\caption{Domain-specific instantiations of Algorithm~\ref{alg:lre}. $B$ is the budget for the
evictable older history $S^\star$ only; in Domain A, the system prompt, task, and last $5$ steps
($R_k$) are concatenated on top without competing for $B$, so the total prompt size is
$\mathrm{tokens}(\text{system}+\text{task}) + B + \mathrm{tokens}(R_k)$ at the cap. In Domain B,
$R_k = \emptyset$ and $B$ is the unit-count cap, so the total equals $B$.}
\label{tab:lre-instantiations}
\end{table}

The scoring stage, value-density ordering, extractive emission, and chronological output are shared
across domains; only $(c_i, B, R_k)$ and the feature map differ. The scorer is a small model over cheap features (\Cref{app:benchmark}) that adds no LLM call
and no neural forward pass at serve time, so the policy's marginal cost is a CPU dot product.
Emission is verbatim: nothing is summarized or rewritten, so exact identifiers, values, and code
survive.

\subsection{Agent Instantiation (Domain A)}
\label{sec:method-agent}

For agentic settings, each unit corresponds to a single $(\text{action}, \text{observation})$ step in the execution trajectory. \method{} represents each unit with a ten-dimensional causal feature vector and an L2-regularized logistic regression; the full feature definitions are given in Appendix Table~\ref{tab:agent-feats}. Retention follows \Cref{alg:app-w} with the Domain A instantiation of Table~\ref{tab:lre-instantiations}.



\subsection{Conversational Instantiation (Domain B)}
\label{sec:method-conv}

For conversational memory, \method{} operates on dialogue-level units, where each unit corresponds to either a dialogue turn or a retrieved session. Across both datasets, \method{} uses standardized trajectory features, fit only on training folds (\Cref{alg:locomo}). With $c_i\equiv 1$ and $R_k=\emptyset$, Equation~\eqref{eq:lre-knapsack} reduces to plain top-$K$ retention with $K=\lceil b\,n \rceil$. Benchmark-specific strategies, deduplication, and labeling are described in \Cref{app:benchmark}.




\subsection{Self-supervised relevance labels}
\label{sec:method-selfsup}

Supervised labels require annotation that a deployment may lack. We therefore derive the label
from the system's own behavior. One principle generates both domain labels: $u_i$ is positive iff
later behavior reuses something $u_i$ contributed,
\begin{equation}
y_i^{\text{self}} \;=\; \mathbb{1}\!\left[\, r(u_i,\, u_{>i}) \,\ge\, \eta \,\right],
\label{eq:selfsup-principle}
\end{equation}
where $r$ is a domain-specific reuse measure and $\eta$ its threshold(\Cref{fig:label-extraction}). 







\begin{table*}[t]\centering\small
\begin{tabular}{@{}l cccc ccc ccc ccc@{}}
\toprule
& \multicolumn{4}{c}{\textbf{Average (168)}} & \multicolumn{3}{c}{\textbf{Easy (57)}}
& \multicolumn{3}{c}{\textbf{Medium (48)}} & \multicolumn{3}{c}{\textbf{Hard (63)}} \\
\cmidrule(lr){2-5}\cmidrule(lr){6-8}\cmidrule(lr){9-11}\cmidrule(lr){12-14}
\textbf{Method} & Acc$\uparrow$ & Stp$\downarrow$ & Pk$\downarrow$ & cL$\downarrow$
& Acc$\uparrow$ & Pk$\downarrow$ & cL$\downarrow$
& Acc$\uparrow$ & Pk$\downarrow$ & cL$\downarrow$
& Acc$\uparrow$ & Pk$\downarrow$ & cL$\downarrow$ \\
\midrule
No compression$^{\dagger}$ & 44.0 & 19.1 & 9.23 & 0.00 & 57.9 & 6.51 & 0.00 & 43.8 & 8.89 & 0.00 & 31.7 & 11.95 & 0.00 \\
\midrule
FIFO (recency)            & 34.5 & 30.6 & 5.65 & 0.00 & 59.6 & 4.98 & 0.00 & 27.1 & 5.20 & 0.00 & \textbf{17.5} & 6.58 & 0.00 \\
ACON (in-fork)            & 26.2 & 33.5 & \textbf{5.12} & 5.45 & 57.9 & \textbf{4.67} & 3.19 & 16.7 & \textbf{4.92} & 5.62 & 4.8 & \textbf{5.68} & 7.37 \\
\rowcolor{mintgreen}
\textbf{\method{} (ours)} & \textbf{41.1} & \textbf{21.8} & 6.91 & \textbf{0.00} & \textbf{73.7} & 5.97 & \textbf{0.00} & \textbf{37.5} & 6.87 & \textbf{0.00} & 14.3 & 7.79 & \textbf{0.00} \\
\bottomrule
\end{tabular}

\caption{AppWorld \texttt{test\_normal} across difficulty levels: Task Goal Completion (Acc, \%), mean
trajectory steps (Stp), peak context tokens in thousands (Pk), and compressor LLM calls per task (cL).
Best deployable value per column in bold; the highlighted row is \textbf{\method{} (ours)}.}
\label{tab:appworld-main}
\end{table*}

\section{Experimental Setup}
\label{sec:setup}

\subsection{Benchmarks}
\label{sec:setup-data}
We evaluate \method{} on AppWorld \citep{trivedi2024appworld}, LoCoMo \citep{maharana2024loco}, and LongMemEval$_S$ \citep{wu2024longmemeval}. AppWorld evaluates long-horizon agent execution under hidden-unit-test evaluation, while LoCoMo and LongMemEval$_S$ evaluate conversational memory retention under query-agnostic and multi-session settings respectively. Detailed benchmark descriptions, splits, labeling strategies, pseudocode and evaluation protocols are reported in Appendix~\ref{app:benchmark}.


\subsection{Baseline taxonomy}
\label{sec:setup-baselines}

We compare against a range of memory retention baselines, including simple heuristics (Recency / FIFO and exponential decay MemoryBank \cite{zhong2024memorybank}), content-based scoring (TF-IDF centroid salience), and dense embedding similarity methods (frozen BGE encoder \cite{xiao2024c}) and LLMLingua-2 \cite{pan2024llmlingua2}. For agentic evaluation, we compare against a no-compression upper bound, FIFO (First-In-First-Out) retention, and ACON (in-fork) \cite{kang2025acon}. ACON demonstrates competitive performance across a range of practical baselines; we therefore include ACON as a representative adaptive in-fork compression method for comparison under the agentic setting. For downstream performance evaluation on AppWorld, LoCoMo, and LongMemEval$_S$, we use GPT-4.1-mini as the base model and GPT-4o as the judge model \cite{hurst2024gpt}. Baseline implementation details, along with pseudocodes are in \Cref{app:baselines}.






\section{Results and Findings}
\label{sec:results}


\subsection{\method{} is Pareto-efficient on long-horizon agents}
\label{sec:results-agent}

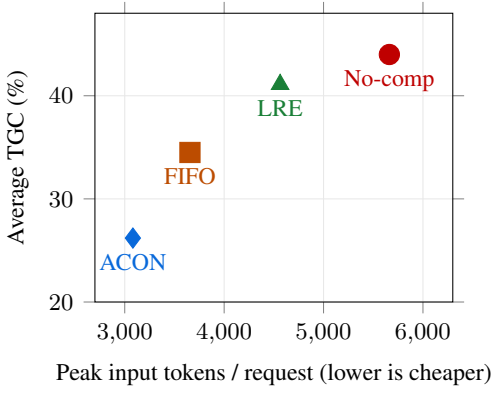
\begin{figure}[t]\centering
\begin{tikzpicture}
\begin{axis}[
  width=0.82\linewidth, height=5.4cm,
  xlabel={Peak input tokens / request (lower is cheaper)},
  ylabel={Average TGC (\%)},
  xmin=2700, xmax=6300, ymin=20, ymax=48,
  grid=both, grid style={gray!18},
  tick align=outside,
  label style={font=\small},
  tick label style={font=\footnotesize}
]

\addplot[
  only marks,
  mark=*,
  mark size=3.8pt,
  color=red!75!black,
  nodes near coords,
  point meta=explicit symbolic,
  every node near coord/.append style={
    anchor=north,
    yshift=-2pt,
    font=\footnotesize
  }
]
coordinates { (5662,44.0) [No-comp] };

\addplot[
  only marks,
  mark=square*,
  mark size=3.8pt,
  color=fifoOrange,
  nodes near coords,
  point meta=explicit symbolic,
  every node near coord/.append style={
    anchor=north,
    yshift=-2pt,
    font=\footnotesize
  }
]
coordinates { (3656,34.5) [FIFO] };

\addplot[
  only marks,
  mark=diamond*,
  mark size=3.8pt,
  color=boxBlue,
  nodes near coords,
  point meta=explicit symbolic,
  every node near coord/.append style={
    anchor=north,
    yshift=-2pt,
    font=\footnotesize
  }
]
coordinates { (3082,26.2) [ACON] };

\addplot[
  only marks,
  mark=triangle*,
  mark size=3.8pt,
  color=lreGreen,
  nodes near coords,
  point meta=explicit symbolic,
  every node near coord/.append style={
    anchor=north,
    yshift=-2pt,
    font=\footnotesize
  }
]
coordinates { (4563,41.1) [LRE] };

\end{axis}
\end{tikzpicture}
\caption{Task success versus peak input tokens on AppWorld;
\method{} attains the highest deployable accuracy at a bounded budget and zero compressor calls}
\label{fig:pareto}
\end{figure}

\method{} achieves task accuracy comparable with retaining full history while operating under a strict context budget (\Cref{tab:appworld-main}, \Cref{fig:pareto}). On Easy tasks, it exceeds the unbounded baseline corresponding to a $+27\%$ relative gain over full-context setting, and remains competitive on Medium tasks where it outperforms both recency ($+38.4\%$) and compression ($+124\%$). This is achieved with zero compressor calls and with a peak prompt that is $25\%$ lower than full context on average, rising to a $35\%$ reduction under Hard settings ($7.79$k vs.\ $11.95$k).

FIFO keeps the budget but discards task-critical state, so the agent loops, spending $+40\%$ more steps per task compared to \method{} (30.6 vs.\ 21.8, and up to 33.5 for ACON). LLM-based compression retains salient content but incurs an inference overhead of $5.45\times$ model calls per task on average (rising to $7.37$ on Hard) while yielding substantially lower task accuracy (if not optimized iteratively). This can be attributed to the loss of execution identifiers during summarization (\S\ref{sec:results-case}).

Matching ACON's recency window to ours ($k{=}5$; $k{=}1$ is its shipped default) recovers accuracy but its peak context rises upto $35.2$k, against \method{}'s $16.5$k at zero compressor calls (\Cref{app:appworld-extra}). We run tests across seed and budget variation, and on two further host models; \method{} holds its accuracy-cost position throughout(\Cref{app:appworld-extra},\Cref{app:crossmodel}, \Cref{tab:crossmodel-gemini}).


\subsection{Why it works: task-critical state survives under extraction}
\label{sec:results-case}

The aggregate numbers leave open why \emph{extraction}, specifically, is what helps. We isolate the
mechanism on a single task (full logs in \Cref{app:case}): a request to build a Spotify playlist from songs the user jotted in Simple Note. Because the task chains two apps, it hinges on several pieces of task-critical state that must all be present when the playlist is assembled. Extractive \method{} ranks the note-content turn and both login turns high and keeps them \emph{verbatim}, so it creates the playlist with the exact token in hand and adds the songs, completing in 19 steps. Recency scrolls the note and the login state out of its window, so the agent re-fetches the note it already read and re-authenticates in a loop, running all the way to the cap without finishing. Abstractive compression preserves the \emph{plan} but also summarizes by paraphrasing and likewise loops to the cap. \method{} thus completes the task, finishes in $37\%$ fewer generation calls than the no-compression (\Cref{fig:growth}). A further example is shown in \Cref{app:case-venmo}. The pattern holds in aggregate; fourteen of the test tasks are properly completed by \method{} alone (\Cref{tab:lre-alone}, \Cref{tab:agent-testfail}).

\subsection{A query-blind scorer that rivals dense retrievers}
\label{sec:results-conv}

\begin{table}[t]\centering\small
\resizebox{\columnwidth}{!}{%
\begin{tabular}{@{}lccc@{}}
\toprule
& \multicolumn{2}{c}{\textbf{Macro-AUC}} & \\
\cmidrule(lr){2-3}
\textbf{Method} & LongMemEval$_S$ & LoCoMo & \textbf{NC}$\downarrow$ \\
\midrule

Recency / FIFO                & 0.527 & 0.478 & 0 \\
MemoryBank (decay)            & 0.527 & 0.478 & 0 \\
Content salience              & \textbf{0.752} & 0.414 & 0 \\

LLMLingua-2                   & 0.402 & --    & 1 \\
Dense salience (BGE, frozen)  & 0.643 & --    & 1 \\

\rowcolor{mintgreen}
\method{}-content             & 0.686 & 0.782 & 0 \\

\rowcolor{mintgreen}
\method{}-self-sup            & 0.698 & 0.769 & 0 \\

\rowcolor{mintgreen}
\textbf{\method{}-supervised} & 0.708 & \textbf{0.829} & 0 \\

\bottomrule
\end{tabular}
}

\caption{Conversational relevance, macro-AUC, deployable (query-blind) methods; NC is neural forward passes/unit. Metric details in \Cref{app:metric}.}
\label{tab:convqa}
\end{table}

The same scorer is a strong ranker on conversational memory, where it must decide what to keep without ever seeing the question (Table~\ref{tab:convqa}). Macro-AUC captures how well a method orders units by relevance: it is the probability that a randomly chosen gold-evidence turn is scored above a randomly chosen non-evidence one, so $1.0$ is a perfect ranking. By this measure, recency and time-decay, are essentially uninformative, because in these long-session dialogue benchmarks, what is relevant is uncorrelated with what is recent. \method{} ranks gold evidence far above them and is the stronger method on LoCoMo. The two baselines that pay a neural forward pass per unit, a frozen dense encoder and a token-pruning compressor, are both outranked by \method{} on LongMemEval$_S$. \method{} uses no forward pass at all: a kilobyte-scale CPU scorer beats GPU encoders.
Overall, an observation is that learned relevance is a stable signal under domain shift, and it can be deployed in practice without any neural passes or expensive inference overhead (Table~\ref{tab:conv-eff}). Gold-recall curves across retention budgets are in \Cref{fig:convqa-recall}.


\subsection{Eviction preserves downstream answer quality}
\label{sec:results-downstream}

Relevance ranking only matters if it improves the final task outcome. \Cref{fig:downstream} in appendix ~\ref{app:conv-extra} evaluates each policy by feeding its retained context to the same QA model under a fixed $20\%$ retention budget. We see that more memory is not always better. On LongMemEval$_S$, retaining the full context produces the worst downstream score (potentially a lost-in-the-middle effect \cite{liu2024lost}). \method{} translates proactive relevance prediction into substantially better budgeted QA performance on LoCoMo. \method{}-supervised achieves the strongest score among deployable policies while reducing context size ($68\%$ fewer tokens than full context) (\Cref{fig:downstream}, \Cref{tab:downstream}). 

\begin{table}[t]\centering\tiny

\begin{tabular}{@{}lcc@{}}
\toprule
\textbf{Metric} & \textbf{LongMemEval$_S$} & \textbf{LoCoMo} \\
\midrule
Self-sup (log label) & 0.698 & 0.769 \\
Supervised (gold) & 0.708 & 0.829 \\
Gap & 0.010 & 0.060 \\
Recovery & \textbf{95\%} & \textbf{82\%} \\
Agent-reuse (transfer) & 0.527 & 0.620 \\
\bottomrule
\end{tabular}
\caption{Self-supervised versus supervised relevance (macro-AUC vs.\ gold). \emph{Recovery of above-chance
skill} is the fraction of the supervised model's discrimination above the $0.5$ chance floor that the
log-derived label retains, $(\mathrm{self\text{-}sup}-0.5)/(\mathrm{supervised}-0.5)$: LongMemEval$_S$
$(0.698-0.5)/(0.708-0.5)=95\%$ and LoCoMo $(0.769-0.5)/(0.829-0.5)=82\%$.}
\label{tab:selfsup}
\end{table}



\subsection{The supervision is already in the logs}
\label{sec:results-selfsup}

We further examine whether self-supervision for \method{} is possible. We replace the human gold label with one the system can produce from its own behavior/log. The log-derived label recovers $95\%$ of the supervised scorer's above-chance discrimination on LongMemEval$_S$ and $82\%$ on LoCoMo (\Cref{tab:selfsup}). We also report a negative transfer result. In agents, a unit is labeled relevant when an identifier or entity it first introduces is later re-referenced by several subsequent units. Such reuse label does not transfer to dialogue, falling near chance on conversational benchmarks. The two domains therefore share a common principle, learning from what the system later used, rather than a single reusable label. The supervision needed for proactive retention is already present in the system's logs.

\section{Conclusion}

We presented LRE, a learned relevance eviction policy that retains
future-useful history through verbatim extraction with a small CPU
scorer. Across both agentic and conversational memory settings, no
evaluated baseline is simultaneously more accurate and cheaper. LRE
approaches full-history performance under a fixed budget, requires no
compressor calls or neural inference, and can be trained from the
system's own later behavior without annotation.

These results suggest that cheap learned relevance is sufficient for
proactive retention. A deployed system can learn what to keep from its
own behavior, preserving task-critical context without neural retention
models or repeated language-model compression.

\section{Limitations}

LRE predicts future utility from signals available at write time. When future dependencies are weakly expressed or absent from the observed trajectory, the scorer can evict information that later becomes important. This limitation is inherent to proactive eviction and applies even with perfect training data. We evaluate on long-horizon agents and conversational memory, but not on multimodal memory, collaborative multi-agent systems, or environments with rapidly changing external state. Whether the same relevance signal transfers to these settings remains an open question. Additional discussion of evaluation, benchmark coverage, feature
dependence, supervision choices, and deployment considerations appears
in Appendix~\ref{app:limitations}.
\clearpage
\bibliography{bib/custom}   
\clearpage
\appendix
\section{Extended Related Work}
\label{app:extended-related}
\paragraph{Anticipating future need.}
The cost of mispredicting need is compute heavy: under recency or LRU, entries are
discarded shortly before reuse and must be recomputed \citep{pan2026kvflow}.
This has motivated \emph{proactive} systems. At the serving level, KVFlow and
PBKV \citep{pan2026kvflow,zheng2026efficient} predict which agent or cache block will fire next from workflow structure and prefetch accordingly; at the agent level, AgentFold \citep{ye2025agentfold} learns to fold and consolidate its own trajectory as it runs. These establish that anticipation beats reaction, but they anticipate \emph{computation} from workflow graphs, or rewrite history through
further LLM calls.  We work along anticipating \emph{which past content} an upcoming query will require, cheaply and without invoking an LLM.

\paragraph{Systems-level context management.}
KV systems such as PagedAttention \citep{kwon2023efficient} remove memory
fragmentation and enable cross-request reuse, raising throughput. Such caching
lowers the cost of \emph{re-processing} a seen context, but doesn't relieve
the cost of \emph{attending} to that context, which becomes error-prone once critical content is buried within a long window \citep{liu2024lost}. 

\paragraph{Self-supervised relevance without labels.}
Training a scorer cheaply requires supervision that costs nothing to obtain.
Hindsight relabeling \citep{andrychowicz2017hindsight} reuses a failed trajectory
by relabeling it with the goal it \emph{did} reach for RL, and contrastive pretraining
\citep{izacard2021unsupervised} learns dense retrievers with no relevance labels at
all. We draw on this idea to derive retention labels directly from what later
turns reused, at no annotation cost.

\section{Benchmark Utility}
\label{app:benchmark}
\subsection{LoCoMo: Query-Agnostic Conversational Eviction}


LoCoMo\cite{maharana2024loco} serves as the primary benchmark for evaluating query-agnostic relevance prediction in conversational context eviction. The dataset contains 5,882 dialogue turns across 10 conversations, where each conversation forms a distinct evaluation group.

\textbf{1. Data structure.}
Each sample corresponds to a dialogue turn $u_i$, and all turns within a conversation form a group $g \in \{1,\dots,10\}$. The task is to assign a relevance score to each turn without conditioning on a specific downstream query. About 24\% of dialogue turns are considered truly relevant (gold evidence across questions)

\textbf{2. Label construction.}
We consider two labeling regimes. In the supervised setting, a turn is labeled positive ($y_i=1$) if it appears in the gold evidence set of any question associated with its conversation. Because these labels are pooled across all the questions tied to the conversation, it yields a query-agnostic relevance signal. The model learns to identify turns that are generally task-critical or informative for the conversation, rather than those that just match a specific query. In the self-supervised setting, a turn is labeled positive if it shares at least 40\% token overlap with an eventual answer for training without explicit annotations. The choice of token overlap percentile is discussed in \Cref{app:justi_conv}

\textbf{3. Splitting and leakage-free training.}
We use a Leave-One-Group-Out (LOGO) protocol over the 10 conversations. In each fold, one full conversation is held out for evaluation while the remaining nine are used for training. To prevent information leakage, the vectorizer and \texttt{StandardScaler} are fitted strictly on training data within each fold. A deduplication guard removes any exact-text overlaps between training and test folds. In LoCoMo, this has minimal effect, removing only 8 training instances corresponding to 3 generic utterances (e.g., “Bye!”, “Take care!”).

\textbf{4. Feature representation and scoring.}
Each unit is mapped to a feature vector combining:
(i) a TF-IDF representation of the text, and
(ii) six trajectory features: position, recency, length, digit count, question-marker presence, and capitalized-word count (\Cref{tab:agent-feats}). A logistic regression model is trained to estimate relevance scores $p_i = P(y_i=1 \mid u_i)$ using these features. At inference time, the model scores all units in the held-out conversation without access to downstream queries.

\textbf{5. Evaluation and downstream usage.}
The model is evaluated both intrinsically and extrinsically. For intrinsic evaluation, we compute ROC-AUC over held-out folds (\Cref{app:metric}). For extrinsic evaluation, retained dialogue turns are passed verbatim into a downstream QA model. Performance is measured using token-level F1, as suggested by \citet{maharana2024loco}.

\begin{figure}[t]
\centering
\fbox{%
\begin{minipage}{0.98\columnwidth}
\label{alg:locomo}
\begin{algorithmic}[1]

\Require LoCoMo dataset $\mathcal{D} = \{(u_i, y_i, g_i)\}$
\Require TF-IDF vectorizer $\phi$, scaler $\sigma$, classifier $f$
\Require Label mode $\mathcal{L} \in \{\text{supervised}, \text{self-supervised}\}$

\vspace{0.5em}
\State \textbf{Dataset structure:}
\State $u_i$: dialogue turn; $g_i$: conversation group

\vspace{0.5em}
\State \textbf{Label construction:}
\State \textit{If supervised:} $y_i = 1$ if $u_i$ appears in gold evidence set
\\
\State \textit{If self-supervised:} $y_i = 1$ if token-overlap$(u_i, \text{answer}) \ge 0.4$

\vspace{0.5em}
\For{each LOGO fold $(\mathcal{D}_{train}, \mathcal{D}_{test})$}

    \State \textbf{Leakage control}
    \State Fit $\phi, \sigma$ on $\mathcal{D}_{train}$ only
    \State Apply transformations to both train and test

    \If{dedup enabled}
        \State Remove exact-text overlaps:
        \State \hspace{1.5em} $\mathcal{D}_{train} \leftarrow \mathcal{D}_{train} \setminus \mathcal{D}_{test}$
    \EndIf

    \vspace{0.5em}
    \State \textbf{Feature construction}

    \State $\mathbf{x}_i^{lex} \leftarrow \phi(u_i)$
    \State $\mathbf{x}_i^{tr} \leftarrow [f_{pos}, f_{rec}, f_{len}, f_{dig}, f_{q}, f_{cap}]$
    
    \State $\tilde{\mathbf{x}}_i^{tr} \leftarrow \sigma(\mathbf{x}_i^{tr})$

    \State $\mathbf{x}_i \leftarrow [\mathbf{x}_i^{lex}; \tilde{\mathbf{x}}_i^{tr}]$

    \vspace{0.5em}
    \State \textbf{Model training}
    \State Train logistic regression $f(\cdot)$ on $(\mathbf{x}_i, y_i)$

    \vspace{0.5em}
    \State \textbf{Inference}
    \State Compute $p_i = f(\mathbf{x}_i)$ for $u_i \in \mathcal{D}_{test}$

\EndFor

\vspace{0.5em}
\State \textbf{Evaluation}
\State ROC-AUC over folds
\State QA evaluation with token-F1 on compressed context

\vspace{0.5em}
\State \Return scorer $f(u_i) \rightarrow p_i$

\end{algorithmic}
\end{minipage}
}
\caption{LoCoMo evaluation pipeline for query-agnostic context eviction in \method{}.}
\end{figure}

\subsection{LongMemEval$_S$: Scalable Session-Level Eviction}

LongMemEval$_S$\cite{wu2024longmemeval} evaluates session-level context eviction under large-scale, query-dependent retrieval settings. Unlike LoCoMo, which operates over dialogue turns, LongMemEval$_S$ defines each eviction unit as a full session within a question-specific haystack.

\textbf{1. Data structure and eviction units.}
The dataset (longmemeval\_s) contains up to 500 questions, where each question defines an independent group. Each group consists of approximately 50 sessions forming a retrieval haystack. The task is highly sparse, with only 5\% of sessions are actually relevant per question (much more sparse / harder retrieval problem).

\textbf{2. Label construction.}

In the supervised setting, a session is labeled positive ($y_i=1$) if its session ID appears in the question-specific \texttt{answer\_session\_ids}. In the self-supervised setting, a proxy label is assigned if the session contains at least 40\% token overlap with the eventual answer for evaluating training performance without annotations (\Cref{app:justi_conv}).

\textbf{3. Splitting and leakage prevention.}
We use grouped K-fold cross-validation with $k=5$ due to the high computational cost of full leave-one-group-out evaluation over 500 questions. A critical challenge in LongMemEval$_S$ is cross-question session reuse. To address this, we apply a deduplication guard that removes any training session whose exact text appears in the held-out question’s haystack. This eliminates over 51,000 duplicate training instances and prevents memorization-based leakage.

\textbf{4. Feature representation and scoring.}
Each session is represented using a fusion of:
(i) TF-IDF lexical features, and
(ii) six standardized trajectory features.

A logistic regression model outputs relevance scores $p_i = P(y_i=1 \mid u_i)$ without access to the downstream question context.

\textbf{5. Evaluation protocol.}
LongMemEval$_S$ is evaluated on both retrieval quality and downstream task performance.For relevance prediction,  we compute ROC-AUC. For downstream evaluation, retained sessions are passed to a QA model (gpt-4.1-mini), and responses are scored using a stronger LLM judge (gpt-4o).
\onecolumn
\begin{figure}[t]
\centering
\fbox{%
\begin{minipage}{0.98\columnwidth}
\label{alg:lme}
\begin{algorithmic}[1]

\Require LongMemEval$_S$ dataset $\mathcal{D} = \{(u_i, y_i, g_i)\}$ with up to 500 questions
\Require TF-IDF vectorizer $\phi$, scaler $\sigma$, classifier $f$
\Require Dedup flag $\texttt{dedup}$

\State \textbf{Dataset structure}
\State Each group $g_i$ is a question with $\sim$50 session-level candidates

\State \textbf{Label construction}
\If{supervised}
    \State $y_i = 1$ if session ID $\in$ answer\_session\_ids$(g_i)$
\Else
    \State $y_i = 1$ if token-overlap$(u_i, \text{answer}) \ge 0.4$
\EndIf

\For{each grouped K-fold split $(\mathcal{D}_{train}, \mathcal{D}_{test})$ with $k=5$}

    \State \textbf{Leakage control}
    \State Fit $\phi, \sigma$ on $\mathcal{D}_{train}$ only

    \If{dedup enabled}
        \State Remove all training sessions whose text appears in $\mathcal{D}_{test}$
    \EndIf

    \State \textbf{Feature construction}
    \State Compute TF-IDF: $x_i^{lex} \leftarrow \phi(u_i)$
    \State Compute trajectory features:
    \State \quad $x_i^{tr} = [f_{pos}, f_{rec}, f_{len}, f_{dig}, f_{q}, f_{cap}]$
    \State Normalize: $\tilde{x}_i^{tr} \leftarrow \sigma(x_i^{tr})$
    \State Concatenate: $x_i \leftarrow [x_i^{lex}; \tilde{x}_i^{tr}]$

    \State \textbf{Training}
    \State Train logistic regression $f(x)$ on $(x_i, y_i)$

    \State \textbf{Inference}
    \State Compute $p_i = f(x_i)$ for test fold sessions

    \State \textbf{Evaluation}
    \State Compute ROC-AUC 
    \State Run downstream QA: gpt-4.1-mini
    \State Score outputs using LLM judge (gpt-4o)

\EndFor

\State \Return scorer $f(u_i) \rightarrow p_i$

\end{algorithmic}
\end{minipage}
}
\caption{LongMemEval$_S$ evaluation pipeline for session-level context eviction under query-dependent retrieval.}
\end{figure}

\twocolumn
\clearpage
\subsection{AppWorld: Agentic Long-Horizon Evaluation}
\label{app:appworld}
\textbf{AppWorld} is a long-horizon tool-use benchmark in which an agent writes Python against application APIs (Venmo, Spotify, Gmail, and others) to complete a natural-language task, graded by hidden unit tests. We evaluate on the \texttt{test\_normal} split (168 tasks: 57 Easy, 48 Medium, 63 Hard), following \cite{kang2025acon} and train the agent scorer on the 90 tasks of the disjoint \texttt{train} split. The agent and, where applicable, the compressor are gpt-4.1-mini.

\textbf{1. Trajectory generation and training data.}
We first execute an unevicted baseline agent ("nocomp") on the AppWorld training split consisting of 90 tasks. This produces full execution trajectories containing interleaved Python code actions and environment observations, which are used as training data for the \method{} scorer.

\textbf{2. Self-supervised labeling via quotation reuse.}
Since no human relevance annotations are available, we construct a zero-supervision label using a hindsight reuse criterion. Let $z_i$ denote an identifier introduced at step $i$. The label is defined as:
\begin{itemize}
    \item $y_i = 1$ if $z_i$ is reused in at least 3 subsequent steps,
    \item otherwise $y_i = 0$.
\end{itemize}
This identifies task-critical identifiers that persist across the execution trajectory.

\textbf{3. Scorer training.}
The logistic regression scorer ($s_{90}$) is evaluated under a Leave-One-Task-Out (LOGO) protocol. 

\textbf{4. Inference and eviction loop.}
During evaluation on test task splits, the agent operates in a ReAct-style loop. When the context length exceeds the token budget ($B = 2048$), the eviction module is triggered. \method{} performs CPU-based scoring of historical units. Units are ranked by score-per-token and retained greedily. This ensures chronological and verbatim reconstruction of the compressed context and preserves exact API identifiers and execution-relevant symbols.

\textbf{5. Evaluation and Task Goal Completion (TGC).}
AppWorld evaluates success using Task Goal Completion (TGC), computed via hidden unit tests that operate on the final environment state.

A known failure mode is \emph{self-reported success bias}, where task success is determined from the agent’s own completion signal rather than from the environment state. This is problematic because the agent can declare success without having actually produced a state that satisfies the underlying task constraints, leading to an empirical inflation of success rates by approximately $2\times$ (\cref{tab:self-report-bias}).

\begin{table}[t]\centering
\tiny
\begin{tabular}{@{}lll@{}}
\toprule
\textbf{Feature} & \textbf{Definition (over prefix $u_{\le i}$)} & \textbf{Domain} \\
\midrule

$f_{\mathrm{pos}}$        & relative position in sequence / trajectory & A, B \\
$f_{\mathrm{loglen}}$     & $\log(1+\text{tokens})$ & A, B \\
$f_{\mathrm{num\_digit}}$ & digit-character count & A, B \\
$f_{\mathrm{caps\_words}}$& capitalized-word count & A, B \\

$f_{\mathrm{recency}}$    & distance to most recent query context & B \\
$f_{\mathrm{has\_q}}$     & presence of question marker / WH pattern & B \\
$f_{\mathrm{tfidf}}$      & TF-IDF lexical representation (10$^4$ cap) & B \\

$f_{\mathrm{len\_tok}}$   & token length & B \\
$f_{\mathrm{has\_error}}$ & error / exception / traceback indicator & A \\
$f_{\mathrm{is\_code}}$   & code-token presence indicator & A \\
$f_{\mathrm{rare\_density}}$ & mean local inverse document frequency & A \\
$f_{\mathrm{sim\_prev}}$  & Jaccard similarity with previous step & A \\
$f_{\mathrm{url\_or\_id}}$& URL / hex$\ge$8 / digit$\ge$4 / handle presence & A\\

\bottomrule
\end{tabular}
\caption{Unified feature space used by \method{} across agentic execution traces (Domain A: AppWorld) and conversational QA (Domain B: LoCoMo and LongMemEval$_S$). The model shares a small set of causal structural features across domains, while augmenting them with domain-specific lexical, conversational, or code-execution signals. All features are computed over prefix histories $u_{\le i}$ to ensure strict causality during training and inference. The feature set is constructed as a minimal causal representation of interaction structure, motivated by observed failure modes across domains.}
\label{tab:agent-feats}
\end{table}

\begin{table}[h]\centering
\tiny
\begin{tabular}{@{}lccc@{}}
\toprule
\textbf{Policy} & \shortstack{Self-reported\\success (\%)} & \shortstack{True TGC\\(replay, \%)} & \shortstack{Inflation\\(self / TGC)} \\
\midrule
No compression            & 98.8 & 44.0 & $2.24\times$ \\
FIFO (recency)            & 60.7 & 34.5 & $1.76\times$ \\
ACON (in-fork)            & 60.1 & 26.2 & $2.30\times$ \\
\textbf{\method{} (ours)} & 89.3 & 41.1 & $2.17\times$ \\
\midrule
\textbf{Pooled (all four)}& \textbf{77.2} & \textbf{36.5} & $\mathbf{2.12\times}$ \\
\bottomrule
\end{tabular}

\caption{Self-reported success versus true TGC (recovered by replaying logged actions into a fresh AppWorld environment and running the hidden unit tests). Across all four policies, self-report overstates success by $1.76$--$2.30\times$ (pooled $2.12\times$), confirming the $\approx 2\times$ inflation and motivating our replay-based evaluation protocol.}
\label{tab:self-report-bias}
\end{table}

To eliminate this source of error, we adopt an evaluation protocol based on deterministic replay. Instead of using any agent-generated success signal, we reconstruct the final environment state from the recorded interaction trace. See the process in \Cref{alg:tgc}.

\onecolumn
\begin{figure}[t]
\centering
\fbox{%
\begin{minipage}{0.98\columnwidth}
\label{alg:app-w}
\begin{algorithmic}[1]

\Require AppWorld environment with 90 training tasks
\Require Agent policy (gpt-4.1-mini), \method{} scorer $f$
\Require Token budget $B = 2048$

\State \textbf{1. Trajectory generation}
\For{each training task $T$}
    \State Run unevicted agent ("nocomp")
    \State Log full trajectory $\tau = \{(a_i, o_i)\}$
\EndFor

\State \textbf{2. Self-supervised labeling}
\For{each step $i$ in trajectory}
    \State Extract identifier $z_i$
    \If{$z_i$ reused $\ge 3$ times in future steps}
        \State $y_i \leftarrow 1$
    \Else
        \State $y_i \leftarrow 0$
    \EndIf
\EndFor

\State \textbf{3. Scorer training}
\State Train logistic regression $s_{90}$ on $(x_i, y_i)$
\State Validate via Leave-One-Task-Out (LOGO), compute AUC

\State \textbf{4. Online agent loop}
\For{each evaluation task}
    \While{agent is executing}
        \State Observe new step $(a_t, o_t)$
        \State Append to history $H$
        
        \If{$\text{tokens}(H) > B$}
            \State Compute scores $p_i = f(u_i)$ for $u_i \in H$
            \State Rank units by $p_i / c_i$
            \State Retain top-scoring units
            \State Emit compressed history verbatim (chronological order)
        \EndIf

        \State Continue ReAct execution (gpt-4.1-mini)
    \EndWhile
\EndFor

\State \textbf{5. Evaluation}
\State Run hidden unit tests to compute TGC

\State \Return Outcome

\end{algorithmic}
\end{minipage}
}
\caption{AppWorld: agentic evaluation of \method{} under token budget constraints and Pareto efficiency criteria.}
\end{figure}

\twocolumn
\onecolumn

\section{Baseline mechanisms and implementation}
\label{app:baselines}
This appendix specifies, for every baseline in both domains, the exact eviction mechanism we implemented, so the comparisons in \S\ref{sec:results} are reproducible. 


\subsection{Agent domain}
\label{app:baselines-agent}

All agent methods share one eviction interface (Algorithm~\ref{alg:agent}): the system prompt, the task, and the last $k$ steps are \emph{always} kept uncapped, and the methods differ only in how the older steps are handled once the history crosses the token budget $B$.
\begin{figure}[t]
\centering
\fbox{%
\begin{minipage}{0.97\columnwidth}
\begin{algorithmic}[1]
\Require History $H=[(a_1,o_1),\dots,(a_t,o_t)]$, budget $B$, keep-last $k$
\State $\mathrm{recent} \gets H[t{-}k{+}1 : t]$ \Comment{last-$k$ steps: always kept, uncapped}
\State $\mathrm{older} \gets H[1 : t{-}k]$
\If{$\mathrm{tokens}(\mathrm{system}) + \mathrm{tokens}(\mathrm{task}) + \mathrm{tokens}(H) \le B$}
    \State \Return $\mathrm{system} \,\Vert\, \mathrm{task} \,\Vert\, H$ \Comment{trigger not reached}
\EndIf
\State $\mathrm{older\_kept} \gets \textsc{HandleOlder}(\mathrm{older}, B)$ \Comment{only per-baseline step}
\State \Return $\mathrm{system} \,\Vert\, \mathrm{task} \,\Vert\, \mathrm{older\_kept} \,\Vert\, \mathrm{recent}$
\Statex
\State \textbf{Per-baseline} $\textsc{HandleOlder}(\mathrm{older},B)$\textbf{:}
\State No-compression: \Return $\mathrm{older}$ \Comment{$B{=}\infty$, never triggers}
\State FIFO: \Return $[\,]$ \Comment{drop older; last-$k$ only}
\State ACON: \Return $[\,\textsc{LlmSummarize}(\mathrm{older})\,]$ \Comment{$+1$ compressor call, lossy}
\State \method{}: \Return $\textsc{LreKnapsack}(\mathrm{older}, B)$ \Comment{verbatim, learned}
\Statex
\Function{LreKnapsack}{$\mathrm{older}, B$}
    \State $\mathrm{units} \gets$ pair each step into one $(a_i,o_i)$ unit
    \For{$u \in \mathrm{units}$}
        \State $p_u \gets f(\mathrm{features}(u))$ \Comment{scorer \texttt{predict\_proba}; 10 causal feats}
    \EndFor
    \State sort $\mathrm{units}$ by $p_u/\mathrm{cost}(u)$ descending \Comment{value density}
    \State $\mathrm{kept} \gets \emptyset,\ \mathrm{used} \gets 0$
    \For{$u \in \mathrm{units}$ in sorted order}
        \If{$\mathrm{used} + \mathrm{cost}(u) \le B$}
            \State $\mathrm{kept} \gets \mathrm{kept} \cup \{u\}$;\ \ $\mathrm{used} \gets \mathrm{used} + \mathrm{cost}(u)$
        \EndIf
    \EndFor
    \State \Return $[\,u \in \mathrm{units} : u \in \mathrm{kept}\,]$ \Comment{verbatim, chronological}
\EndFunction
\end{algorithmic}
\end{minipage}}
\caption{Domain A (agent): shared eviction trigger; methods differ only in \textsc{HandleOlder}.
No-compression keeps everything (unbounded), FIFO drops the older steps, ACON replaces them with one abstractive LLM summary, and \method{} keeps the highest learned-relevance-per-token steps verbatim.}
\label{alg:agent}
\end{figure}
\twocolumn
\paragraph{Notes.} No-compression is the ``keep everything'' upper bound (prompt grows with the horizon). FIFO is verbatim but importance-blind: any task-critical fact past the last-$k$ window is lost. ACON adds one compressor LLM call and uses ACON's native $\mathrm{preserve\_last\_k}{=}1$ (the default in \cite{kang2025acon}), whereas FIFO and \method{} use $k{=}5$. \method{} is bounded, importance-aware, verbatim, and language-model-free. The scorer is a learner over causal features (Table~\ref{tab:agent-feats}).

\paragraph{ACON compression prompt.} We use ACON's shipped (not iteratively optimized) compression prompt verbatim (the system prompt and the \texttt{prompt\_history\_v2}), changing only the budget to $B{=}2048$. The code for ACON was kept unmodified and used just as shipped\footnote{Forked from \url{https://github.com/microsoft/acon} at commit \texttt{d63f9ae}.}.

\begin{qualbox}{ACON compression SYSTEM prompt}
\begin{lstlisting}[style=snip]
# ACON compression SYSTEM prompt (context_opt/system_prompt.jinja):
You are an agent tasked with extracting and refining a concise and optimized
version of the context based on the user instruction and other provided information.

# ACON history-compression GUIDELINE (context_opt/prompt_history_v2.jinja):
You are maintaining a structured context-aware summary for a productivity agent. You
will be given the user instruction for the agent, a list of interactions corresponding
to actions taken by the agent, and the most recent previous summary if one exists.
Produce the following:

### REASONING
Summarize key progress, decisions made, important observed outcomes, and rationale
behind actions taken so far. Include how earlier steps influenced later ones and why
certain data is retained in the summary.

### COMPLETED
List completed subtasks or successful outcomes, with brief results if applicable.

## [Information Source]
### USER INSTRUCTION
{{ task }}
## [PREVIOUS SUMMARY] (if any)
{{ prev_summary }}
## [HISTORY OF INTERACTIONS]
{{ history }}

## PRIORITIZE
1. Keep all sections relevant and concise.
2. Use reusable structured formats when summarizing artifacts.
3. Ensure agent can resume task with no loss of information.
4. Include key info from errors or failed attempts to prevent repeated mistakes.
5. Preserve all essential artifacts and data needed to complete the task.

### [Output Format]
Do not include the input or any additional explanation. Only return the formatted summary.
\end{lstlisting}
\end{qualbox}

\subsection{Conversational domain}
\label{app:baselines-conv}

Every conversational method reduces to a per-unit \emph{keep-score} (higher means retain) over the units of one group (a conversation in LoCoMo, a question's haystack in LongMemEval$_S$); retention keeps the top budget-fraction. Algorithm~\ref{alg:conv} gives the shared loop and every baseline's score.

\begin{figure}[t]
\centering
\fbox{%
\begin{minipage}{0.97\columnwidth}
\begin{algorithmic}[1]
\Require Group of units $U=\{u_1,\dots,u_n\}$, score function $\mathrm{score}(\cdot)$, budget fraction $b$
\Require $\mathrm{score}$ is query-blind or reads the question (reference)
\State \textbf{Scoring}
\For{each unit $u_i \in U$}
    \State $s_i \gets \mathrm{score}(u_i, U)$ \Comment{per-baseline; cases below}
\EndFor
\State \textbf{Retention}
\State $K \gets \lceil b\,n \rceil$
\State $\mathrm{keep} \gets$ indices of the $K$ highest $s_i$
\State \Return $[\,u_i : i \in \mathrm{sort}(\mathrm{keep})\,]$ \Comment{verbatim, chronological}
\Statex
\State \textbf{Per-baseline} $\mathrm{score}(u_i,U)$\textbf{:}
\State Recency: $s_i \gets 1/(1+\mathrm{age}_i)$
\State MemoryBank: $s_i \gets \exp(-10\,\mathrm{age}_i/(n{-}1))$ \Comment{monotone in age $\Rightarrow$ recency}
\State Content salience: $s_i \gets \cos\!\big(\mathrm{tfidf}(u_i),\bar{c}_U\big)$ \Comment{centroid typicality}
\State Dense salience: $s_i \gets \cos\!\big(\mathrm{emb}(u_i),\bar{e}_U\big)$ \Comment{1 encoder pass / unit}
\State LLMLingua-2: $s_i \gets \mathrm{mean}_{t \in u_i} P(\mathrm{keep}\mid t)$ \Comment{calibrated keep-class}
\State \method{}: $s_i \gets P(\mathrm{relevant}\mid u_i)$ \Comment{logistic regression; no neural pass}
\end{algorithmic}
\end{minipage}}
\caption{Domain B (conversational): shared score-then-retain eviction, with each baseline's per-unit score.
$\mathrm{age}_i$ is distance from the most recent unit; $\bar{c}_U,\bar{e}_U$ are the group TF-IDF / embedding
centroids; query-aware references read the question and are upper bounds, never competitors.}
\label{alg:conv}
\end{figure}

\paragraph{Notes.}
MemoryBank~\cite{zhong2024memorybank} is the Ebbinghaus curve $R=\exp(-t/S)$ with $S{=}1,\,t{=}\mathrm{age}$; with no access stream it is strictly age-monotone and therefore induces the \emph{same ranking as recency}. Content and dense salience score typicality relative to the group centroid, not relevance to any query (query blind). LLMLingua-2~\cite{pan2024llmlingua2} runs the pretrained \texttt{llmlingua-2-bert} token classifier and is task-agnostic (it never reads a question); its generic \texttt{LABEL\_0/1} keep-class is calibrated empirically against information-dense probes. Dense embeddings use a frozen BGE encoder\footnote{BGE model: https://huggingface.co/BAAI/bge-small-en-v1.5}.

\paragraph{\method{} variants.} All three share one model (an L2 logistic regression with balanced class weights, vectorizer and scaler fit per fold on training units only) and the inference
$s_i=P(\mathrm{relevant}\mid u_i)$ with no neural pass. They differ only in label and features:
\begin{itemize}\itemsep1pt
\item \textbf{\method{}-supervised}. Label: human gold evidence already given in the benchmark.
\item \textbf{\method{}-self-sup} (zero-annotation). Label: the response-overlap proxy (a turn is positive if its content tokens cover $\ge 0.4$ of some eventual answer's), derived from later responses with no annotation needed. \Cref{app:justi_conv} shows the justification for the choice of token coverage value.
\item \textbf{\method{}-content} (ablation). Label: human gold evidence (same as \method{}-supervised). Features: the TF-IDF text vector \emph{alone}, with the six trajectory features removed. This isolates the contribution of the trajectory block: the gap between \method{}-supervised and \method{}-content on a given dataset is what those six features add over text-only scoring.

\end{itemize}

\section{Justification of Self-Supervised Label Design Choices}
\label{app:justify_label}

We evaluate label quality using two quantities:
(i) \textbf{AUC-vs-gold}, which measures how well self-supervised labels align with human-annotated gold relevance labels (higher is better; 0.5 corresponds to random alignment), and
(ii) \textbf{positive rate}, which denotes the fraction of units labeled as positive under a given rule. For agentic settings, we additionally report \textbf{LOTO AUC} (Leave-One-Task-Out AUC), which measures how well a classifier trained on these labels generalizes to unseen tasks.

\subsection{Conversational Overlap Threshold ($\tau = 0.4$)}
\label{app:justi_conv}
For conversational self-supervision, a turn is labeled positive if its content overlaps with a downstream response above a threshold $\tau$. We select $\tau = 0.4$ based on a joint sweep across LoCoMo and LongMemEval$_S$, evaluating AUC-vs-gold and label sparsity.

Table~\ref{tab:overlap_sweep} summarizes the results.

\begin{table}[h]
\centering
\tiny
\setlength{\tabcolsep}{3pt}
\begin{tabular}{c|cccc}
\toprule
$\tau$ & LoCoMo AUC & LongMemEval$_S$ AUC & LoCoMo+ & LongMemEval$_S$+ \\
\midrule
0.2 & 0.74 & 0.65 & 0.94 & 0.47 \\
0.3 & 0.75 & 0.69 & 0.85 & 0.38 \\
\textbf{0.4} & \textbf{0.77} & \textbf{0.70} & 0.66 & 0.29 \\
0.5 & 0.76 & 0.68 & 0.61 & 0.22 \\
0.6 & 0.76 & 0.68 & 0.39 & 0.13 \\
0.8 & 0.75 & 0.65 & 0.30 & 0.11 \\
\bottomrule
\end{tabular}
\caption{Sweep over response-overlap threshold $\tau$.}
\label{tab:overlap_sweep}
\end{table}

\textbf{Why $\tau=0.4$ is chosen:}
First, it is the \emph{joint maximizer} of AUC-vs-gold across both datasets. Second, it lies in a flat optimum region ($\tau \in [0.3,0.5]$), where performance varies by at most $0.016$ AUC. This indicates stability.

\textbf{Why not other values:}
Lower thresholds ($\tau=0.2, 0.3$) produce excessively high positive rates on LoCoMo ($0.85$--$0.94$). This collapses the ranking problem into near-uniform positives and weakens discrimination. Such a near-degenerate label distribution does not reflect real-world deployment nuances. Higher thresholds ($\tau \ge 0.5$) reduce coverage too aggressively (down to $0.11$--$0.61$ positive rate) and degrade alignment with gold annotations.

\subsection{Agentic Reuse Breadth ($b = 3$)}
\label{app:justi_agent}
For agentic trajectories, a step is labeled positive if an identifier it introduces is reused at least $b$ times in subsequent steps. We evaluate $b \in \{1,2,3,4,5\}$ using Leave-One-Task-Out evaluation.

\begin{table}[h]
\centering
\small
\begin{tabular}{c|ccc}
\toprule
$b$ & Pos-rate & \# Positives & LOTO AUC \\
\midrule
1 & 0.811 & 1269 & 0.878 \\
2 & 0.582 & 911 & 0.830 \\
\textbf{3} & \textbf{0.432} & \textbf{676} & \textbf{0.828} \\
4 & 0.335 & 524 & 0.835 \\
5 & 0.271 & 424 & 0.847 \\
\bottomrule
\end{tabular}
\caption{Sweep over reuse breadth $b$ for agentic self-supervision.}
\label{tab:breadth_sweep}
\end{table}

\textbf{LOTO AUC} (Leave-One-Task-Out Area Under the ROC Curve) measures how well a classifier trained on these labels ranks truly relevant steps above irrelevant ones in unseen tasks. A value of 0.5 corresponds to random ranking, while 1.0 corresponds to perfect separation.

\textbf{Why $b=3$ is chosen:}
Unlike the conversational case, LOTO AUC is relatively flat across all values ($0.83$--$0.88$). We therefore select $b=3$ based on semantic and operational criteria.

First, $b=1$ is overly permissive, marking 81\% of steps as positive. This collapses the eviction task into a near-majority-class problem, where ranking becomes dominated by trivial structural correlations rather than task-critical utility. Second, $b=3$ yields a moderate positive rate ($\approx 0.43$), which aligns with the expected compression domain \cite{kang2025acon} in which only a subset of trajectory state can be retained.

\textbf{Why not other values:}
Values $b \ge 4$ become too selective, reducing the positive set to $27$--$34\%$, which harms coverage of reusable context. $b \le 2$ over-labels transient identifiers that are only briefly reused. We choose a parameter value that reflects real deployment nuances and therefore fix $b=3$, a semantic choice that captures repeatedly-reused, task-critical state while maintaining a stable and informative training signal.
\section{Metrics: definitions and recovery}
\label{app:metric}

This appendix defines every reported score in both domains, then describes how the agent metric and its costs are recovered honestly.

\subsection{Conversational relevance scores}

\paragraph{ROC-AUC (per group).} Each method assigns a real-valued keep-score $s_i$ to every unit $u_i$ of a group, scored against the binary gold label $y_i\in\{0,1\}$. Let $\mathcal{P}=\{i:y_i{=}1\}$ and $\mathcal{N}=\{i:y_i{=}0\}$ be the relevant and irrelevant units, with $P=|\mathcal{P}|$, $N=|\mathcal{N}|$. The area under the receiver-operating-characteristic curve equals the probability that a random relevant unit
outscores a random irrelevant one, computed exactly as the normalized Mann--Whitney statistic \cite{mann1947test}
{\small
\[
\mathrm{AUC}\;=\;\frac{1}{P\,N}\sum_{i\in\mathcal{P}}\sum_{j\in\mathcal{N}}
\Big(\mathbf{1}[\,s_i>s_j\,]+\tfrac{1}{2}\,\mathbf{1}[\,s_i=s_j\,]\Big).
\]
}
$\mathrm{AUC}=1$ is a perfect ranking (every relevant unit above every irrelevant one), $0.5$ is chance
(random ordering), and a value $<0.5$ means the score is \emph{anti}-correlated with relevance. AUC is
threshold-free and invariant to the positive rate, so it measures the ranking quality an eviction policy needs (what to keep first) without
committing to a budget. A group is scored only if it contains at least one relevant and one irrelevant unit
(otherwise AUC is undefined).

\paragraph{Macro-AUC.} The per-group AUCs are averaged with equal weight,
$\text{macro-AUC}=\frac{1}{|G|}\sum_{g\in G}\mathrm{AUC}_g$, so a large group cannot dominate a small one; the groups are conversations on LoCoMo and questions on LongMemEval$_S$.

\paragraph{Recall@budget.} A retention policy keeps the $K=\lceil b\,n\rceil$ highest-scoring of the $n$ units at budget fraction $b\in\{0.1,0.2,0.3,0.4\}$. Recall@$b=|\text{kept}\cap\mathcal{P}|/P$ is the fraction of
relevant units that survive, averaged over groups.

\paragraph{Budget@recall (iso-accuracy efficiency).} The inverse statistic: the smallest budget fraction $b$
at which recall@$b$ reaches a target (e.g.\ $0.8$), averaged over groups, reported as the saving relative to
recency at the same recall.

\subsection{Conversational answer-quality scores (downstream)}

\paragraph{Token-F1 (LoCoMo).} With the answerer reading only the retained context, we compare its answer $A$ to the gold answer $G$ as bags of normalized content tokens. With $\mathrm{prec}=|A\cap G|/|A|$ and
$\mathrm{rec}=|A\cap G|/|G|$,
\[
\text{token-F1}=\frac{2\,\mathrm{prec}\cdot\mathrm{rec}}{\mathrm{prec}+\mathrm{rec}},
\]
the standard extractive-QA overlap, averaged over questions.

\paragraph{LLM-judge (LongMemEval$_S$).} Free-form answers are graded by a model \emph{stronger} than the answerer (gpt-4o judging gpt-4.1-mini, to avoid self-grading): the judge sees the question, the gold answer, and the candidate, and returns a binary correct/incorrect vote; the score is the fraction judged correct. The judge was validated on crafted correct, wrong, and paraphrase cases.

\subsection{Conversational serving-cost scores}

\paragraph{Throughput, neural calls, dollar cost.} \emph{Units per second} is the measured rate at which a
loaded scorer scores resident units (hardware recorded). \emph{Neural forward passes
per unit} counts the encoder or classifier passes a method performs per scored unit.

\subsection{Agent scores}

\paragraph{Task Goal Completion (TGC).} The all-or-nothing success metric on AppWorld: a task scores $1$ only
if \emph{every} hidden unit test passes (partial credit such as $3$ of $4$ tests counts as $0$), and $\text{TGC}\%$ over a difficulty bucket is the mean over its tasks. How we recover true TGC (rather than the inflated self-report) is in \S\ref{app:metric-recovery}.


\paragraph{Cost axes.} \emph{cLLM} is compressor language-model calls per task (the number of compression events). \emph{Input tokens per request} (in/req) is the mean prompt size per generation, our peak-token proxy since the true per-step peak is not logged. \emph{Total tokens per task} sums agent and recovered
compressor tokens. 


\subsection{Recovering the agent metric and costs}
\label{app:metric-recovery}

\paragraph{Recovering true TGC by replay.} The agent framework ships with its environment evaluation disabled, so the run's self-reported completion flag overstates success and is not Task Goal Completion. We recover true TGC at zero model cost: because a freshly constructed environment resets task state, we replay each
task's logged action sequence into a fresh environment and then call the environment's evaluation, which passes only when all hidden unit tests pass. The replay reconstructs the final state deterministically rather than reporting the empty-environment artifact. All reported TGC values use this recovered metric, never the self-report. Algorithm~\ref{alg:tgc} gives the procedure.

\begin{figure}[t]
\centering
\fbox{%
\begin{minipage}{0.97\columnwidth}
\begin{algorithmic}[1]
\Require Task id $T$, logged action sequence (the agent's code blocks)
\State $\mathrm{world} \gets \textsc{AppWorld}(T)$ \Comment{fresh env: resets to $T$'s initial state}
\For{each $\mathrm{action}$ in the logged sequence}
    \State $\mathrm{world}.\mathrm{execute}\big(\mathrm{clean\_code}(\mathrm{action})\big)$ \Comment{replay to rebuild final state; errors swallowed}
\EndFor
\State $r \gets \mathrm{world}.\mathrm{evaluate}()$ \Comment{run hidden unit tests on the reconstructed state}
\State $\mathrm{tgc} \gets \big(r.\mathrm{num\_passes} = r.\mathrm{num\_tests}\big) \wedge \big(r.\mathrm{num\_failures} = 0\big)$ \Comment{all-or-nothing}
\State \Return $\mathrm{tgc}$ \Comment{written to \texttt{tgc.json}; never the agent self-report}
\end{algorithmic}
\end{minipage}}
\caption{Recovery of true TGC by replay (\$0, no LLM call): \texttt{replay\_eval.py}. A fresh
environment resets to the initial state, so we replay the logged actions to reconstruct the agent's final
state before evaluating.}
\label{alg:tgc}
\end{figure}



\section{Additional AppWorld Results}
\label{app:appworld-extra}

\subsection{Peak Prompt Tokens}
Table~\ref{tab:peak-range} reports per-task peak prompt statistics on AppWorld \texttt{test\_normal}. This table shows the corresponding minimum, maximum, mean, and median values. LRE has the lowest maximum peak among the evaluated policies. FIFO and ACON keep a bounded working context, but failed runs can still reach larger task-level peaks because they continue to the step cap.


\begin{table}[H]\centering
\tiny
\begin{tabular}{@{}lccccc@{}}
\toprule
\textbf{Method} & Min peak & Max peak & Mean & Median & \shortstack{Peak saving\\(vs worst max)} \\
\midrule
No compression            & 3{,}250 & 25{,}043 & 9{,}230 & 7{,}592 & 21.5\% \\
FIFO (recency)            & 3{,}287 & 28{,}565 & 5{,}646 & 5{,}155 & 10.5\% \\
ACON (in-fork)            & 3{,}643 & 31{,}918 & 5{,}121 & 4{,}832 & --- \\
\textbf{\method{} (ours)} & 3{,}545 & \textbf{15{,}388} & 6{,}912 & 6{,}640 & \textbf{52\%} \\
\bottomrule
\end{tabular}


\caption{Per-task peak prompt tokens on AppWorld \texttt{test\_normal}. FIFO and ACON loop to the step cap,
so their max peak \emph{exceeds} full context; \method{} is the only bounded method whose max peak falls
below it, saving as much as $52\%$ against the worst-case peak (ACON's).}
\label{tab:peak-range}
\end{table}

\subsection{Hidden Unit-Test Failures}
Table~\ref{tab:agent-testfail} reports hidden unit-test failure rates. The pattern is similar to Task Goal Completion: FIFO and ACON fail substantially more tests overall, while LRE remains closer to the no-compression baseline.

\begin{table}[H]\centering\small
\begin{tabular}{@{}l cccc@{}}
\toprule
& \multicolumn{4}{c}{\textbf{Unit-test failure rate (\%, $\downarrow$)}} \\
\cmidrule(lr){2-5}
\textbf{Method} & Avg & Easy & Med & Hard \\
\midrule
No compression$^{\dagger}$ & 27.0 & 17.6 & 22.3 & 33.9 \\
\midrule
FIFO (recency)            & 51.6 & 18.1 & 52.7 & 64.8 \\
ACON (in-fork)            & 58.7 & 19.9 & 58.3 & 74.9 \\
\textbf{\method{} (ours)} & \textbf{36.0} & \textbf{8.8} & \textbf{28.3} & \textbf{52.2} \\
\bottomrule
\end{tabular}

\caption{Unit-test failure rate on AppWorld \texttt{test\_normal} by difficulty: the fraction of hidden unit tests that fail (lower is better), the partial-credit complement to the all-or-nothing TGC of
Table~\ref{tab:appworld-main}. Tests per split: Easy $216$, Medium $336$, Hard $525$ ($1{,}077$ total)}
\label{tab:agent-testfail}
\end{table}

\subsection{Tasks solved only by LRE}

Table~\ref{tab:lre-alone} lists the AppWorld \texttt{test\_normal} tasks solved only by LRE.

\begin{table}[H]\centering\small
\begin{tabular}{@{}llcccc@{}}
\toprule
\textbf{Task} & \textbf{Diff.} & No-comp & FIFO & ACON & \textbf{\method{}} \\
\midrule
\texttt{024c982\_2} & E & 6/7 & 0/7 & 6/7 & \textbf{7/7} \\
\texttt{09b0ee6\_1} & E & 1/2 & 1/2 & 1/2 & \textbf{2/2} \\
\texttt{552869a\_2} & E & 1/2 & 1/2 & 1/2 & \textbf{2/2} \\
\texttt{59fae45\_3} & E & 2/6 & 4/6 & 5/6 & \textbf{6/6} \\
\midrule
\texttt{0d01c76\_3} & M & 2/6 & 1/6 & 1/6 & \textbf{6/6} \\
\texttt{270f1ff\_3} & M & 1/11 & 0/11 & 1/11 & \textbf{11/11} \\
\texttt{3aa1a22\_1} & M & 6/7 & 2/7 & 2/7 & \textbf{7/7} \\
\texttt{3d9a636\_1} & M & 2/5 & 0/5 & 1/5 & \textbf{5/5} \\
\texttt{d194965\_2} & M & 5/6 & 3/6 & 3/6 & \textbf{6/6} \\
\texttt{d6ac34d\_2} & M & 8/9 & 7/9 & 2/9 & \textbf{9/9} \\
\texttt{ff58e36\_1} & M & 2/6 & 1/6 & 1/6 & \textbf{6/6} \\
\midrule
\texttt{2c544f9\_3} & H & 2/6 & 1/6 & 2/6 & \textbf{6/6} \\
\texttt{9016950\_3} & H & 3/7 & 3/7 & 2/7 & \textbf{7/7} \\
\texttt{f323bae\_2} & H & 1/9 & 0/9 & 1/9 & \textbf{9/9} \\
\bottomrule
\end{tabular}

\caption{The $14$ AppWorld \texttt{test\_normal} tasks (seed 1) solved by \method{} alone: hidden unit-test
pass counts (passed/total) per policy. \method{} passes \emph{all} tests on every task, while
no-compression, FIFO, and ACON each leave at least one test failing.}
\label{tab:lre-alone}
\end{table}

\subsection{Recency-matched ACON ($k{=}5$)}
\label{app:acon-k5}

Matching the recency window recovers accuracy but its worst-case peak rises to $32.2$k (Medium) and $35.2$k (Hard). Against ACON as a competitive baseline, \method{} achieves comparable gains at substantially lower cost (\Cref{tab:acon-k5}). 

\begin{table*}[t]\centering\small
\begin{tabular}{@{}l cc cc cc@{}}
\toprule
& \multicolumn{2}{c}{\textbf{TGC (\%)}$\uparrow$}
& \multicolumn{2}{c}{\textbf{max peak (k tok)}$\downarrow$}
& \multicolumn{2}{c}{\textbf{compressor calls}$\downarrow$} \\
\cmidrule(lr){2-3}\cmidrule(lr){4-5}\cmidrule(lr){6-7}
\textbf{Method} & Med & Hard & Med & Hard & Med & Hard \\
\midrule
FIFO ($k{=}5$)                  & 27.1 & 17.5 & \textbf{8.6} & 30.1 & 0.0 & 0.0 \\
ACON $k{=}1$ (shipped default)  & 16.7 & \phantom{0}4.8 & 9.0 & 34.7 & 5.6 & 7.4 \\
ACON $k{=}5$ (recency-matched)  & 35.4 & 12.7 & 32.2 & 35.2 & 2.9 & 4.1 \\
\rowcolor{mintgreen}
\textbf{\method{} (ours)}       & \textbf{37.5} & 14.3 & 10.4 & \textbf{16.5} & \textbf{0.0} & \textbf{0.0} \\
\bottomrule
\end{tabular}

\caption{Recency-matched ACON on AppWorld Medium ($n{=}48$) and Hard ($n{=}63$). \emph{max peak} is the per-task maximum prompt size (gpt-4.1-mini).}
\label{tab:acon-k5}
\end{table*}
\begin{table*}[t]\centering\small
\begin{tabular}{@{}l ccc c c@{}}
\toprule
& \multicolumn{3}{c}{\textbf{per seed}} & & \\
\cmidrule(lr){2-4}
\textbf{Arm} & s1 & s2 & s3 & \textbf{mean $\pm$ SE} & \textbf{max peak range} \\
\midrule
\multicolumn{6}{@{}l}{\emph{TGC (\%)}} \\
ACON ($k{=}5$)            & 35.4 & 27.1 & 33.3 & $31.9 \pm 2.5$ & --- \\
\rowcolor{mintgreen}
\textbf{\method{} (ours)} & 37.5 & 39.6 & 33.3 & $\mathbf{36.8 \pm 1.9}$ & --- \\
\midrule
\multicolumn{6}{@{}l}{\emph{max peak (k tokens)}} \\
ACON ($k{=}5$)            & 32.2 & 33.3 & 25.6 & $30.4 \pm 2.4$ & $[25.6,\,33.3]$ \\
\rowcolor{mintgreen}
\textbf{\method{} (ours)} & 10.4 & \textbf{10.1} & 19.4 & $\mathbf{13.3 \pm 3.0}$ & $\mathbf{[10.1,\,19.4]}$ \\
\bottomrule
\end{tabular}

\caption{Three-seed variance on AppWorld Medium ($n{=}48$). Accuracy ties within noise; the worst-case peak ranges are disjoint.}
\label{tab:seeds}
\end{table*}

\subsection{Seed variance}
\label{app:seeds}

\Cref{tab:seeds} reports three seeds of the key Medium comparison.

\subsection{Budget sensitivity}
\label{app:budget-sweep}

\Cref{tab:budget-sweep} sweeps a $3\times$ range. TGC rises
monotonically with budget, as expected, but no pairwise difference is significant (McNemar $p{=}0.14$ for $B{=}1024$ vs.\ $B{=}3072$).

\begin{table}[H]\centering\small
\begin{tabular}{@{}l c ccc@{}}
\toprule
& & \multicolumn{3}{c}{\textbf{peak prompt (k tok)}$\downarrow$} \\
\cmidrule(lr){3-5}
\textbf{Budget $B$} & \textbf{TGC (\%)}$\uparrow$ & median & p90 & max \\
\midrule
1024 ($\tfrac{1}{2}\times$)  & 27.1 & \textbf{5.9} & \textbf{8.0} & 27.0 \\
\rowcolor{mintgreen}
2048 (paper)                 & 37.5 & 6.7 & 8.8 & \textbf{10.4} \\
3072 ($1.5\times$)           & \textbf{41.7} & 7.3 & 9.2 & 10.7 \\
\bottomrule
\end{tabular}

\caption{Budget sweep for \method{} on AppWorld Medium.}
\label{tab:budget-sweep}
\end{table}
\section{Cross-Model Transfer}
\label{app:crossmodel}

The scorer is external to the host model and consumes only logged text, so in principle it should transfer across model families. We test that directly on two further hosts, Gemini-2.5-Flash (Google) and DeepSeek-V3, on AppWorld Medium.

\subsection{Gemini-2.5-Flash}

We apply the gpt-4.1-mini scorer unchanged (\emph{zero-shot}) and, additionally re-fit a scorer on Gemini's own logged trajectories (\Cref{tab:crossmodel-gemini}).

\begin{table*}[t]\centering\small
\begin{tabular}{@{}l c l c@{}}
\toprule
\textbf{Arm} & \textbf{TGC (\%)}$\uparrow$ & \textbf{95\% CI} & \textbf{max peak (k tok)}$\downarrow$ \\
\midrule
No compression (ceiling) & 45.8 & $[32.6,\ 59.7]$ & 46.9 \\
\midrule
\rowcolor{mintgreen}
\textbf{\method{} (gpt scorer, zero-shot)} & \textbf{29.2} & $[18.2,\ 43.2]$ & 14.8 \\
\rowcolor{mintgreen}
\textbf{\method{} (Gemini scorer, re-fit)} & \textbf{29.2} & $[18.2,\ 43.2]$ & 14.8 \\
FIFO (recency)                             & 10.4 & $[4.5,\ 22.2]$ & \textbf{9.8} \\
\bottomrule
\end{tabular}

\caption{Gemini-2.5-Flash host, AppWorld Medium ($n{=}48$), Wilson $95\%$ intervals.}
\label{tab:crossmodel-gemini}
\end{table*}

\begin{table*}[t]\centering\small
\begin{tabular}{@{}l l c c c@{}}
\toprule
\textbf{\method{} scorer} & \textbf{trained on} & \textbf{LOTO AUC} & \textbf{TGC (\%)} & \textbf{tasks resolved identically} \\
\midrule
gpt-4.1-mini (zero-shot) & gpt-4.1-mini logs & 0.828 & 29.2 & \multirow{2}{*}{44/48 (92\%)} \\
Gemini-native (re-fit)   & Gemini logs       & \textbf{0.864} & 29.2 & \\
\bottomrule
\end{tabular}

\caption{Zero-shot versus host-native scorer on the Gemini host. Higher in-domain AUC,
identical downstream outcome.}
\label{tab:crossmodel-scorer}
\end{table*}

\subsection{DeepSeek-V3}

DeepSeek-V3's true ceiling on this split is low ($22.9\%$) for a reason unrelated to memory; it over-reports completion, self-reporting success on $98$--$100\%$ of tasks against $\approx 23\%$ true TGC under replay evaluation (\Cref{tab:crossmodel-deepseek}). This is the same self-report bias documented
in \Cref{tab:self-report-bias}. Its failures are premature completion
rather than dropped state, so eviction policy has little to act on and every arm ties.

\begin{table*}[t]\centering\small
\begin{tabular}{@{}l c c c@{}}
\toprule
\textbf{Arm} & \textbf{TGC (\%)} & \textbf{max peak (k tok)}$\downarrow$ & \textbf{compressor calls}$\downarrow$ \\
\midrule
No compression                  & 22.9 & 14.8 & 0.00 \\
ACON (\textbf{gpt} compressor)  & 22.9 & 24.5 & 1.58 \\
\rowcolor{mintgreen}
\textbf{\method{} (ours)}       & 22.9 & \textbf{9.9} & \textbf{0.00} \\
\bottomrule
\end{tabular}

\caption{DeepSeek-V3 host, AppWorld Medium ($n{=}48$). The ACON arm uses a \emph{gpt}
compressor, a stronger-compressor control. All arms tie on accuracy; \method{} holds the
lowest peak at zero compressor calls.}
\label{tab:crossmodel-deepseek}
\end{table*}

Read together, where failures are
memory-bound, transfer holds and the margin is at least as large as in-domain; where they are
not, \method{} costs nothing and still bounds the peak.
\section{Additional Conversational Results}
\label{app:conv-extra}

\subsection{Retention and Downstream Performance}

Figure~\ref{fig:downstream} reports downstream QA performance under fixed retention budgets.


On LongMemEval$_S$, the deployable policies cluster more closely and content salience remains slightly ahead of \method{}.Content salience leads on this benchmark, because its topically distinctive sessions make centroid typicality a strong proxy for relevance but that advantage is \emph{local}, not general: the same method drops to near chance on LoCoMo, where \method{} dominates. Even in this harder setting, however, \method{} remains competitive.

\begin{figure*}[t]\centering
\begin{tikzpicture}
\begin{groupplot}[
  group style={group size=2 by 1, horizontal sep=1.3cm},
  width=0.45\linewidth, height=5.6cm,
  xlabel={Context tokens / question ($\times10^3$, $\downarrow$)},
  grid=both, grid style={gray!18}, tick align=outside,
  label style={font=\small}, title style={font=\small\bfseries},
  tick label style={font=\footnotesize},
  every axis plot/.append style={line width=0.9pt, mark size=4.5pt}]

\nextgroupplot[title={LoCoMo (token-F1)}, ylabel={Answer token-F1 ($\uparrow$)},
  xmin=1, xmax=21, ymin=0, ymax=0.26, ytick={0,0.05,0.10,0.15,0.20,0.25},
  legend style={font=\scriptsize, draw=none, fill=white, fill opacity=0.75, text opacity=1,
    at={(0.97,0.04)}, anchor=south east}, legend cell align=left]
\addplot[only marks, mark=o, line width=1.2pt, color=black!55] coordinates {(18.8,0.233)};
  \addlegendentry{Full context$^{\dagger}$}
\addplot[only marks, mark=square*, color=fifoOrange] coordinates {(3.6,0.100)};
  \addlegendentry{Recency}
\addplot[only marks, mark=diamond*, color=boxBlue] coordinates {(4.0,0.081)};
  \addlegendentry{Content salience}
\addplot[only marks, mark=triangle*, color=lreGreen] coordinates {(6.5,0.160)};
  \addlegendentry{\method{}-self-sup}
\addplot[only marks, mark=star, color=lreGreen, mark size=3pt] coordinates {(6.0,0.177)};
  \addlegendentry{\textbf{\method{}-supervised}}

\nextgroupplot[title={LongMemEval\_S (LLM-judge)}, ylabel={Judge correctness ($\uparrow$)},
  xmin=23, xmax=53, ymin=0, ymax=0.40, ytick={0,0.1,0.2,0.3,0.4}]
\addplot[only marks, mark=o, line width=1.2pt, color=black!55] coordinates {(50.0,0.242)};
\addplot[only marks, mark=square*, color=fifoOrange] coordinates {(26.1,0.276)};
\addplot[only marks, mark=diamond*, color=boxBlue] coordinates {(36.5,0.338)};
\addplot[only marks, mark=triangle*, color=lreGreen] coordinates {(38.9,0.280)};
\addplot[only marks, mark=star, color=lreGreen, mark size=3pt] coordinates {(38.8,0.296)};
\end{groupplot}
\end{tikzpicture}
\caption{Downstream answer quality versus token cost under a $20\%$ retention budget (Full context
$\dagger$ = no eviction). The comparison set is minimal and deployment-oriented: recency is the production
default, content salience the strongest non-neural query-blind competitor, and the \method{} variants test
whether proactive relevance transfers to QA; MemoryBank (ranking-equivalent to recency) and the neural
salience baselines (dominated by content salience) are omitted.}
\label{fig:downstream}
\end{figure*}
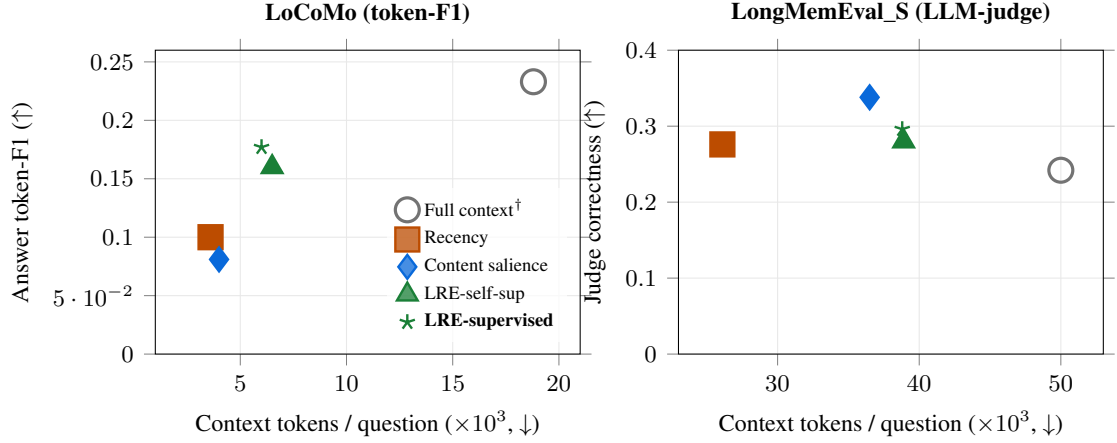

\begin{figure*}[t]\centering
\begin{tikzpicture}
\begin{groupplot}[
  group style={group size=2 by 1, horizontal sep=1.3cm},
  width=0.45\linewidth, height=5.4cm,
  xlabel={Retention budget (\% kept)},
  xmin=8, xmax=42, ymin=0, ymax=90,
  grid=both, grid style={gray!18}, tick align=outside,
  label style={font=\small}, title style={font=\small\bfseries},
  tick label style={font=\footnotesize},
  every axis plot/.append style={thick, mark size=3.2pt}]

\nextgroupplot[title={LoCoMo (token-F1 gold)}, ylabel={Gold recall (\%)},
  legend style={font=\scriptsize, draw=none, fill=white, fill opacity=0.75, text opacity=1,
    at={(0.03,0.97)}, anchor=north west}, legend cell align=left]
\addplot[color=fifoOrange, mark=square*] coordinates {(10,9.2) (20,19.0) (40,39.3)};
  \addlegendentry{Recency}
\addplot[color=boxBlue, mark=diamond*] coordinates {(10,4.7) (20,11.1) (40,31.5)};
  \addlegendentry{Content salience}
\addplot[color=lreGreen, mark=triangle*] coordinates {(10,24.8) (20,43.9) (40,70.4)};
  \addlegendentry{\method{}-self-sup}
\addplot[color=lreGreen, mark=star, mark size=4.8pt] coordinates {(10,29.3) (20,50.8) (40,78.6)};
  \addlegendentry{\textbf{\method{}-supervised}}

\nextgroupplot[title={LongMemEval\_S}, ylabel={Gold recall (\%)}]
\addplot[color=fifoOrange, mark=square*] coordinates {(10,13.9) (20,25.3) (40,45.0)};
\addplot[color=boxBlue, mark=diamond*] coordinates {(10,36.8) (20,53.5) (40,75.2)};
\addplot[color=lreGreen, mark=triangle*] coordinates {(10,20.8) (20,38.8) (40,69.6)};
\addplot[color=lreGreen, mark=star, mark size=4.8pt] coordinates {(10,22.4) (20,39.5) (40,70.7)};
\end{groupplot}
\end{tikzpicture}
\caption{Gold recall versus retention budget on both benchmarks. On LoCoMo, \method{} retains more gold evidence than every baseline at every budget; on LongMemEval$_S$ \method{} tracks just below content salience but reaches it at a higher budget.}
\label{fig:convqa-recall}
\end{figure*}
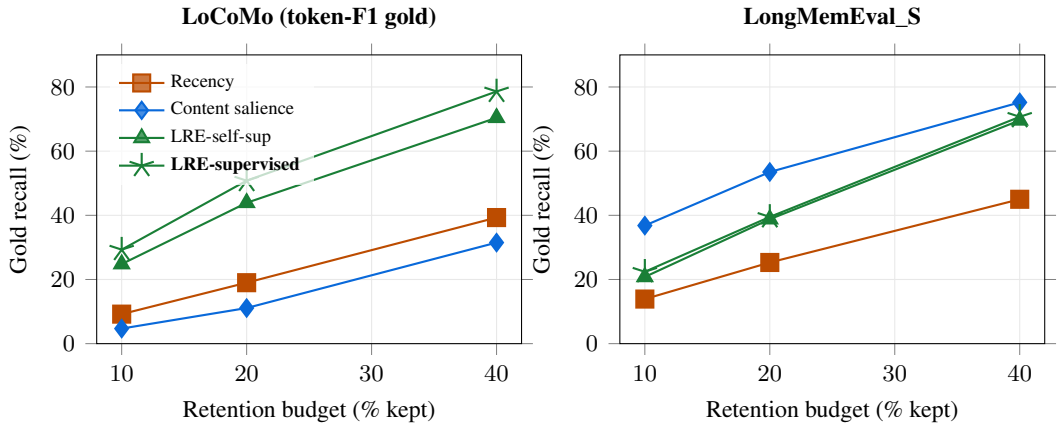

\begin{table*}[t]\centering\small
\begin{tabular}{@{}l cc cc@{}}
\toprule
& \multicolumn{2}{c}{\textbf{LoCoMo (token-F1)}} & \multicolumn{2}{c}{\textbf{LongMemEval$_S$ (LLM-judge)}} \\
\cmidrule(lr){2-3}\cmidrule(lr){4-5}
\textbf{Policy} & Acc$\uparrow$ & ctx tok$\downarrow$ & Acc$\uparrow$ & ctx tok$\downarrow$ \\
\midrule
Full context$^{\dagger}$      & 0.233 & 18.8k & 0.242 & 50.0k \\
\midrule
Recency                       & 0.100 & \textbf{3.6k} & 0.276 & \textbf{26.1k} \\
Content salience              & 0.081 & 4.0k & \textbf{0.338} & 36.5k \\

\rowcolor{mintgreen}
\method{}-self-sup            & 0.160 & 6.5k & 0.280 & 38.9k \\

\rowcolor{mintgreen}
\textbf{\method{}-supervised} & \textbf{0.177} & 6.0k & 0.296 & 38.8k \\
\bottomrule
\end{tabular}
\caption{Downstream answer quality at a $20\%$ retention budget. On LoCoMo (token-F1) the full context
is best but token-heavy, and \method{} leads the budget-constrained policies; on LongMemEval$_S$
(LLM-judge) the full context is worst and the deployable policies cluster.}
\label{tab:downstream}
\end{table*}

\subsection{Serving Cost and Self-Supervision}
Table~\ref{tab:conv-eff} reports serving cost and iso-recall budget savings. Table~\ref{tab:selfsup} compares the self-supervised and supervised variants. Hardware: LoCoMo on Apple M2 (8-core arm64); LongMemEval$_S$ on Modal Linux (8 vCPU x86\_64 + NVIDIA T4 16 GB). LRE methods run CPU-only on both; dense and LLMLingua use the T4 on LongMemEval$_S$. Within-dataset throughput ratios are hardware-controlled; cross-dataset comparisons should not be made because unit sizes differ.

\begin{table*}[t]\centering\small
\begin{tabular}{@{}llccc@{}}
\toprule
\textbf{Data} & \textbf{Method} & \shortstack{neural\\calls/unit} & \shortstack{units/sec\\(CPU)} & \shortstack{budget@80\%\\(saved vs rec.)} \\
\midrule
LoCoMo & Recency                & 0 & 7{,}444{,}187 & 0.834 (--) \\
LoCoMo & Content salience       & 0 & \phantom{0,000}57{,}152 & 0.839 ($-0.6\%$) \\
LoCoMo & \textbf{\method{}-sup} & \textbf{0} & \textbf{129{,}081} & \textbf{0.411 ($50.8\%$)} \\
\midrule
LongMemEval$_S$ & Recency & 0 & 2{,}836{,}670 & 0.622 (--) \\LongMemEval$_S$ & \textbf{\method{}-sup} & \textbf{0} & 1{,}181 & \textbf{0.364 ($41.5\%$)} \\
LongMemEval$_S$ & Dense salience (BGE) & 1 & \phantom{1{,}}103 & 0.449 ($27.9\%$) \\
LongMemEval$_S$ & LLMLingua-2   & 1 & \phantom{1{,0}}36 & 0.690 ($-10.9\%$) \\
\bottomrule
\end{tabular}

\caption{Conversational serving cost and iso-recall budget saving. \method{} does zero neural forward passes
and scores resident units on a CPU, whereas the dense and LLMLingua-2 encoders pay one forward pass per unit. At iso-recall ($80\%$ gold recall) \method{} reaches the target at roughly half
the budget of recency.}
\label{tab:conv-eff}
\end{table*}


\subsection{Scorer Size}
\label{app:scorer-size}

\method{}
is $295\times$ smaller than the dense encoder and $1569\times$ smaller than the token-pruning
compressor, and beats both on LongMemEval$_S$ ($0.708$ vs.\ $0.643$ and $0.402$), at zero
neural forward passes and zero language-model calls (\Cref{tab:scorer-size}).

\begin{table*}[t]\centering\small
\begin{tabular}{@{}l l r r r c@{}}
\toprule
\textbf{Method} & \textbf{Type} & \textbf{Params} & \textbf{Weights} & $\times$\textbf{vs conv} & \textbf{AUC (LoCoMo / LME$_S$)} \\
\midrule
\rowcolor{mintgreen}
\textbf{\method{} agent scorer} & logistic (10 feats) & ${\sim}10$ & \textbf{1.7\,KB} & --- & --- \\
\rowcolor{mintgreen}
\textbf{\method{} conv scorer}  & TF-IDF + LR & $\le 10^4$ & \textbf{226 / 441\,KB} & $1\times$ & \textbf{0.829 / 0.708} \\
\midrule
BGE-small-en-v1.5      & dense encoder      & 33\,M  & 127\,MB & $295\times$  & --- / 0.643 \\
LLMLingua-2 (mBERT)    & token-pruning LM   & 177\,M & 677\,MB & $1569\times$ & --- / 0.402 \\
\bottomrule
\end{tabular}

\caption{Serialized scorer size versus ranking quality. \method{} AUC is reported as
LoCoMo / LongMemEval$_S$; the neural baselines were run on LongMemEval$_S$ only and their
weights are benchmark-independent.}
\label{tab:scorer-size}
\end{table*}
\section{Extended limitations}
\label{app:limitations}

This appendix expands the limitations stated in the main text. We group them by scope: evaluation, method, supervision,
measurement, and generalization. Each limitation lists what we did to mitigate it and, where relevant, what a deployment would need to verify before relying on \method{}.

\paragraph{In-fork ACON baseline.} ACON's published fork ships the
unoptimized compression prompt and we used it as it is. The fair comparison that licenses our claim is the matched in-fork one (identical agent, identical seed, identical evaluation harness).

\paragraph{Benchmark coverage.} We evaluate on three benchmarks: LoCoMo and LongMemEval$_S$ for
conversational memory, AppWorld for long-horizon agents. These cover dialogue and tool-use coding
agents but do not include web-browsing agents, multimodal agents, or open-domain coding agents (e.g., SWE-bench). The Pareto-efficiency claim is supported within these regimes.

\paragraph{Subset of AppWorld used.} We evaluate on \texttt{test\_normal} (168 tasks). We do not
evaluate on \texttt{test\_challenge} (417 tasks).

\paragraph{Budget-bounded performance.} On
tasks whose required state exceeds the operating budget (AppWorld Hard at $B=2048$ being the
clearest case) no-compression wins because retaining everything is the only way to keep state
that does not fit. \method{} therefore complements, rather than replaces, unbounded retention when
the deployment can afford it. The right deployment question is whether the budget is binding; where it is, \method{} is Pareto-efficient.

\paragraph{Eviction unit granularity.} Our agent unit is the (action, observation) pair; our
conversational unit is the dialogue turn or session. We do not evict at the token level inside a
single forward pass, which is what KV-cache methods such as
H\textsubscript{2}O \citep{zhang2023h2o}, SnapKV \citep{li2024snapkv}, and TRIM-KV \citep{bui2025trimkv} do. Token-level methods are
complementary: they operate inside the model and require model-internal signals; \method{}
operates outside the model and requires only logged behavior. A deployment could plausibly combine
them (\method{} for unit-level retention, token pruning inside each retained unit), but we do not
evaluate that combination here.

\paragraph{Feature-set dependence.} The learned scorer operates on a compact set of causal features that we designed to be inexpensive, model-agnostic, and available at decision time. While these features transfer across the benchmarks studied here, we do not claim that they are universally optimal. Different deployments may expose different signals, interaction patterns, memory units, or notions of future utility. In practice, a production system should treat our feature set as a reference implementation rather than a fixed specification, and should measure and validate its own candidate features from logged behavior before deployment. The broader claim of \method{} is that future utility can be learned from causal signals available at write time, not that any particular feature set is universally sufficient.

\paragraph{Fixed self-supervised threshold.} The conversational self-supervised label uses a
fixed answer-recall threshold of $\tau = 0.4$, jointly optimal on LoCoMo and LongMemEval$_S$. In deployment, however, the right rule is to set $\tau$ implicitly (through a
target positive rate on a sample of logged exchanges) rather than to inherit our value. The $\tau = 0.4$ choice is justified empirically (Table~\ref{tab:overlap_sweep}); we do not claim
it is universal.

\paragraph{No multimodal or extended unit types.} The eviction unit is text. Multimodal agents
whose task-critical state includes images, audio, or large code blobs would require the unit
representation and the token-cost function to be extended; we have not done so.

\paragraph{What the reuse proxy can miss.} The self-supervised label defines relevance as
``later reused,'' which is a measure of the eviction objective rather than of importance in
the abstract. Under this label, content that is important but never restated, referenced, or
acted upon again is negative. A constraint stated once and then honored implicitly is the
clearest at-risk case, and a scorer trained only on this signal could in principle learn to
evict that whole class. However, the choice of features and system requirements ultimately
belongs to the practitioner. The practitioner can therefore encode signals beyond reuse
when the application requires them, and we argue that \method{} should be understood as a
lightweight, configurable eviction mechanism rather than a universal measure of importance.
\clearpage
\onecolumn

\section{Qualitative case study: Venmo request}
\label{app:case-venmo}

This appendix gives the full qualitative example summarized in \S\ref{sec:results-case}: an outcome
(Example~A) and the controlled mechanism (Example~B), both drawn from AppWorld task
\texttt{024c982\_2}. 

\begin{qualbox}{Qualitative example A: only \method{} completes the task}
\textbf{Task} (\texttt{024c982\_2}): \emph{``Request \$28 privately on Venmo from my roommate,
Melissa, with a note, `For the movie tickets'.''}

\smallskip
\begin{center}\tiny
\begin{tabular}{@{}lccl@{}}
\toprule
\textbf{Policy} & \textbf{Steps} & \textbf{TGC} & \textbf{Failure mode} \\
\midrule
No compression          & 13       & \textcolor{aconRed}{\ding{55}} & self-completes but fails 1 of 7 hidden tests (6/7) \\
FIFO (recency)          & 49 (cap) & \textcolor{aconRed}{\ding{55}} & loses login state $\to$ re-authenticates in a loop (33/49 login steps); 0/7 \\
ACON (LLM compression)  & 49 (cap) & \textcolor{aconRed}{\ding{55}} & summary drops the token $\to$ re-authenticates (34/49 login steps); self-reports done but 6/7 \\
\textbf{\method{} (ours)}     & \textbf{18} & \textcolor{lreGreen}{\textbf{\ding{51}}} & completes correctly, all 7 tests pass (7/7) \\
\bottomrule
\end{tabular}
\end{center}

\end{qualbox}

\begin{qualbox}{Qualitative example B: why the task-critical credential survives only under \method{}}
The task hinges on one \textbf{task-critical identifier}: the Venmo \texttt{access\_token} obtained at
login, required by \emph{every} subsequent API call. How each policy treats it under the same
$B{=}2048$ budget:

\smallskip
\textbf{\textcolor{lreGreen}{\method{} (extractive)}}: the scorer ranks the login turn high and keeps it
\emph{verbatim} within budget, so the exact token is still present at step~17, which creates the
request directly:
\begin{lstlisting}[style=snip]
# step 17: I will now create a private payment request of $28 to her
#          with the note "For the movie tickets" using the access token.
payment_request = apis.venmo.create_payment_request(access_token=..., amount=28,
                  description="For the movie tickets", private=True)
-> {"message": "Payment request created.", "payment_request_id": 6097}   # then complete_task()
\end{lstlisting}

\textbf{\textcolor{fifoOrange}{FIFO (recency)}}: \texttt{preserve\_last\_k=5} scrolls the login
turn out as history grows, so the agent ``forgets'' it already authenticated and re-logs-in. It spends
\textbf{33 of 49} steps on login/token actions before hitting the step cap:
\begin{lstlisting}[style=snip]
# (repeated) I have the Venmo password. I will now login to Venmo using ...
-> {"access_token": "eyJhbGciOiJIUzI1NiIsInR5cCI6IkpXVCJ9...."}   # re-issued, never used to finish
\end{lstlisting}

\textbf{\textcolor{aconRed}{ACON (abstractive)}}: its LLM summary preserves the \emph{reasoning}
but \textbf{drops the exact token} (the lossy-compression failure mode). It too re-authenticates
(\textbf{34 of 49} steps); it eventually self-reports completion but still fails a hidden test:
\begin{lstlisting}[style=snip]
### REASONING  The agent's objective is to request $28 privately on Venmo from
Melissa with the note "For the movie tickets." ... identified the correct API
endpoint (create_payment_request) ... retrieved the Venmo password from stored
credentials ...            # keeps "$28", "Melissa", "create_payment_request";
                           # NO access_token  ->  the agent must re-authenticate
\end{lstlisting}

\smallskip
\textbf{Takeaway.} The task turns on an exact opaque credential. \textbf{Extractive} eviction (\method{})
preserves it bit-for-bit; \textbf{recency} (FIFO) scrolls it out; \textbf{abstractive compression}
(ACON) paraphrases it away. Only \method{} still holds the value the final action depends on.
\end{qualbox}

\section{Qualitative case study: Simple Note playlist}
\label{app:case}

This appendix gives the full qualitative example summarized in \S\ref{sec:results-case}: an outcome
(Example~A) and the controlled mechanism (Example~B), both drawn from AppWorld task
\texttt{d194965\_2}. 

\begin{qualbox}{Qualitative example A: only \method{} completes the task}
\textbf{Task} (\texttt{d194965\_2}): \emph{``I jotted down some songs in Simple Note recently. Make a
playlist titled `Songs from Simple Note' out of it.''}

\smallskip
\begin{center}\tiny
\begin{tabular}{@{}lccl@{}}
\toprule
\textbf{Policy} & \textbf{Steps} & \textbf{TGC} & \textbf{Failure mode} \\
\midrule
No compression          & 30       & \textcolor{aconRed}{\ding{55}} & self-completes but fails 1 of 6 hidden tests (5/6) \\
FIFO (recency)          & 49 (cap) & \textcolor{aconRed}{\ding{55}} & loses the note and login state $\to$ re-fetches and re-authenticates in a loop (34/49 login steps); 3/6 \\
ACON (LLM compression)  & 49 (cap) & \textcolor{aconRed}{\ding{55}} & summary drops the tokens and song list $\to$ re-authenticates (31/49 login steps); 3/6 \\
\textbf{\method{} (ours)}     & \textbf{19} & \textcolor{lreGreen}{\textbf{\ding{51}}} & completes correctly, all 6 tests pass (6/6) \\
\bottomrule
\end{tabular}
\end{center}

\end{qualbox}

\bigskip

\begin{qualbox}{Qualitative example B: why the task-critical state survives only under \method{}}
This task chains two apps and a data hand-off, so it hinges on several \textbf{task-critical
identifiers} that must all be in context at the playlist-assembly step: the Simple Note
\texttt{access\_token}, the \textbf{song list} extracted from the note, the Spotify
\texttt{access\_token}, and the \texttt{playlist\_id}. How each policy treats them under the same
$B{=}2048$ budget:

\smallskip
\textbf{\textcolor{lreGreen}{\method{} (extractive)}}: the scorer ranks the note-content turn and both login
turns high and keeps them \emph{verbatim}, so at step~13 the exact Spotify token is still present and the
playlist is created directly (then its songs are added):
\begin{lstlisting}[style=snip]
# step 13: logged in to Spotify; create the playlist using this access token.
spotify_access_token = spotify_login_response["access_token"]
create_playlist_response = apis.spotify.create_playlist(access_token=spotify_access_token,
                           title="Songs from Simple Note")
-> {"message": "Playlist created.", "playlist_id": 654}
\end{lstlisting}

\smallskip
\textbf{\textcolor{fifoOrange}{FIFO (recency)}}: \texttt{preserve\_last\_k=5} scrolls the note content
and login state out as history grows, so the agent ``forgets'' both --- it re-fetches the note it
already read and re-logs-in, spending \textbf{34 of 49} steps on login/token actions before the cap:
\begin{lstlisting}[style=snip]
# (repeated) I have the content of the note which contains the songs. I will parse ...
# (repeated) The Spotify login API requires both username and password. I will try to login ...
\end{lstlisting}

\smallskip
\textbf{\textcolor{aconRed}{ACON (abstractive)}}: its LLM summary preserves the \emph{plan} (read the
note, build the playlist) but \textbf{drops the exact tokens and the song list} (the lossy-compression
failure mode). It too re-authenticates (\textbf{31 of 49} steps) and never assembles all the state at
once, hitting the cap and failing:
\begin{lstlisting}[style=snip]
### REASONING  The agent's objective is to create a Spotify playlist titled "Songs
from Simple Note" from songs noted in Simple Note ... located the relevant note ...
logged in to retrieve credentials ...   # keeps the plan and titles;
                                        # NO access_token, NO verbatim song list
\end{lstlisting}

\smallskip
\textbf{Takeaway.} The task turns on exact opaque state (two access tokens, the playlist id, and the verbatim song list) that a paraphrase cannot reconstruct. \textbf{Extractive} eviction (\method{}) preserves it bit-for-bit; \textbf{recency} (FIFO) scrolls it out; \textbf{abstractive compression} (ACON) paraphrases it away. Only \method{} still holds the values the final actions depend on.
\end{qualbox}

\twocolumn

\end{document}